\documentclass[runningheads]{llncs}
\usepackage[T1]{fontenc}
\usepackage{algorithm}
\usepackage{algorithmic}
\usepackage{amsmath}
\usepackage{multirow}
\usepackage{amssymb}
\usepackage{graphicx}
\usepackage{multirow}

\usepackage[misc]{ifsym}

\usepackage{booktabs}
\usepackage{caption}
\usepackage{pifont}
\usepackage{makecell}
\usepackage{comment}
\usepackage{tikz}
\usepackage{rotating} 
\usepackage{array} 
\usepackage{orcidlink}
\begin{document}
\title{RetinexDual: Retinex-based Dual Nature Approach for Generalized Ultra-High-Definition Image Restoration}
\titlerunning{RetinexDual}
%
\author{Mohab Kishawy \inst{(\textrm{\Letter})}\orcidlink{0000-0003-0679-7068} \and
Ali Abdellatif Hussein\orcidlink{0009-0002-5084-3420} \and
Jun Chen\orcidlink{0000-0002-8084-9332}}
\authorrunning{M. Kishawy et al.}
%
\institute{Department of Electrical \& Computer Engineering, McMaster University, Hamilton, Ontario, Canada \\
\email{kishawym@mcmaster.ca}}
\maketitle              
\begin{abstract}
Advancements in image sensing have elevated the importance of Ultra-High-Definition Image Restoration (UHD IR). Traditional methods, such as extreme downsampling or transformation from the spatial to the frequency domain, encounter significant drawbacks: downsampling induces irreversible information loss in UHD images, while our frequency analysis reveals that pure frequency-domain approaches are ineffective for spatially confined image artifacts, primarily due to the loss of degradation locality. To overcome these limitations, we present RetinexDual, a novel Retinex theory-based framework designed for generalized UHD IR tasks. RetinexDual leverages two complementary sub-networks: the Scale-Attentive maMBA (SAMBA) and the Frequency Illumination Adaptor (FIA). SAMBA, responsible for correcting the reflectance component, utilizes a coarse-to-fine mechanism to overcome the causal modeling of mamba, which effectively reduces artifacts and restores intricate details. On the other hand, FIA ensures precise correction of color and illumination distortions by operating in the frequency domain and leveraging the global context provided by it. Evaluating RetinexDual on four UHD IR tasks, namely deraining, deblurring, dehazing, and Low-Light Image Enhancement (LLIE), shows that it outperforms recent methods qualitatively and quantitatively. Ablation studies demonstrate the importance of employing distinct designs for each branch in RetinexDual, as well as the effectiveness of its various components.

\end{abstract}

\section{Introduction}

Recently, the capabilities of imaging sensors and displays have been evolving to provide detailed and sophisticated Ultra-High-definition (UHD) images. Despite these advancements, reversing the distortion and degradation of the captured information caused by light exposure, haze, motion, or weather conditions has become more challenging. This paper addresses these challenges to achieve a unified UHD Image Restoration (IR) framework.


    


Due to the breakthrough in deep learning methods, learning-based approaches and architectures were proposed to address IR challenge \cite{lv2024fourier,9710307,cai2023retinexformer}, attaining state-of-the-art performance on general image restoration. However, these approaches were not able to handle the UHD image sizes and the extensive information in them due to the computational power needed. This limitation has emerged as a significant hurdle to UHD image restoration.

Aiming to improve UHD IR, research studies were conducted specifically for this purpose \cite{wang2024uhdformer,Li2023ICLR,yu2022towards,Zhao_2025_CVPR}. Wang et al. \cite{wang2024uhdformer} proposed in UHDformer the potential of using a correction transformer that utilizes high-resolution features to attend to low-resolution restoration. On the other hand, UHDFour \cite{Li2023ICLR} suggested processing the image in the frequency domain after extreme downsampling to enhance lighting conditions. However, following such a "Downsampling-Enhancement-Upsampling" modeling paradigm, crucial details are removed due to the separation between resampling operations and the enhancement process \cite{Yu_2024_CVPR}. Furthermore, Wave-Mamba \cite{zou2024wavemamba}, and ERR \cite{Zhao_2025_CVPR} focused on restoring UHD images by operating on various frequency bands to leverage the global context and the simplified analysis of certain frequency-based patterns achieved in the frequency domain.


\begin{figure}[t]
  \centering
  \setlength{\tabcolsep}{1pt} 
  
  \begin{minipage}{0.49\textwidth}
    \centering
    \begin{tabular}{cccc}
      \multirow{2}{*}{\rotatebox{90}{Dehaze}} & 
      \raisebox{0.3\height}{\rotatebox{90}{Input}} & 
      \includegraphics[width=0.43\columnwidth]{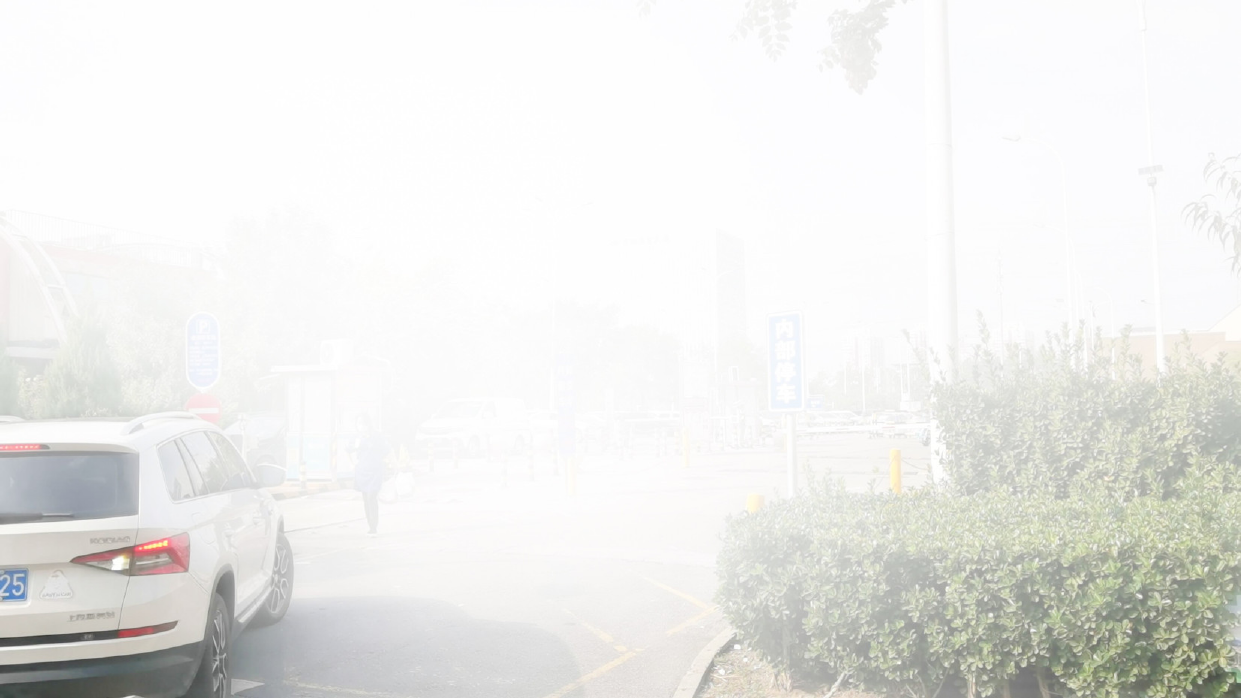} &
      \includegraphics[width=0.43\columnwidth]{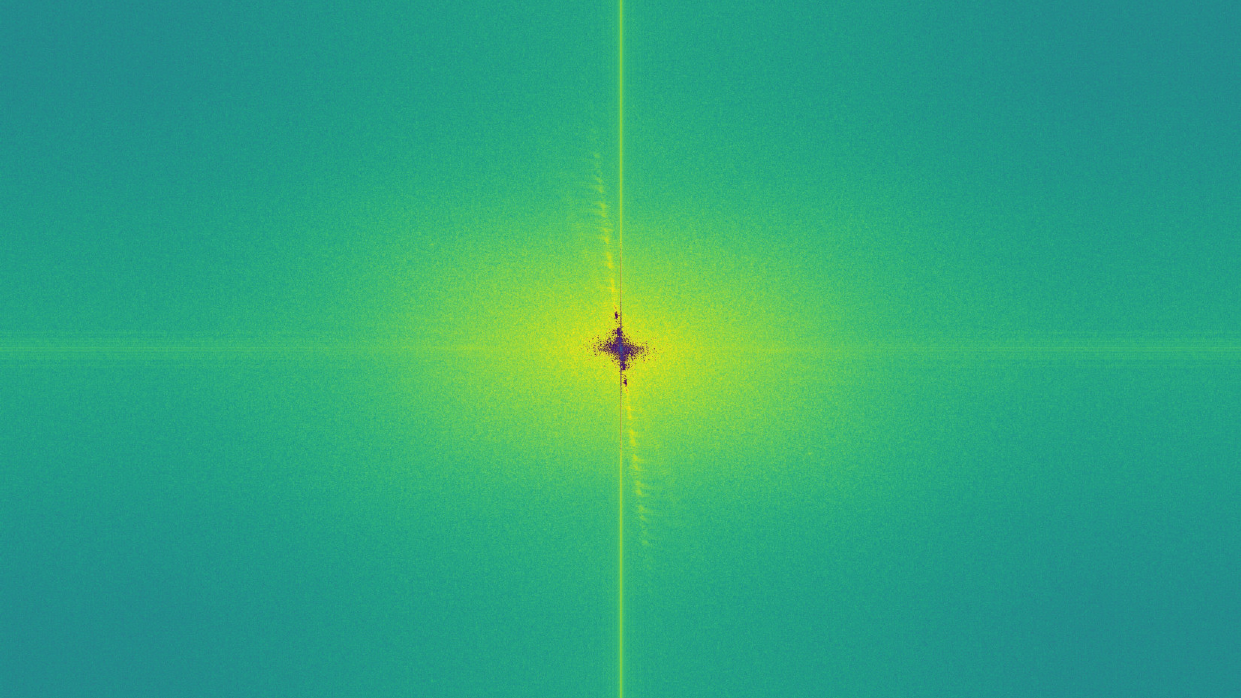} \\
      &\raisebox{\height}{\rotatebox{90}{GT}} & 
      \includegraphics[width=0.43\columnwidth]{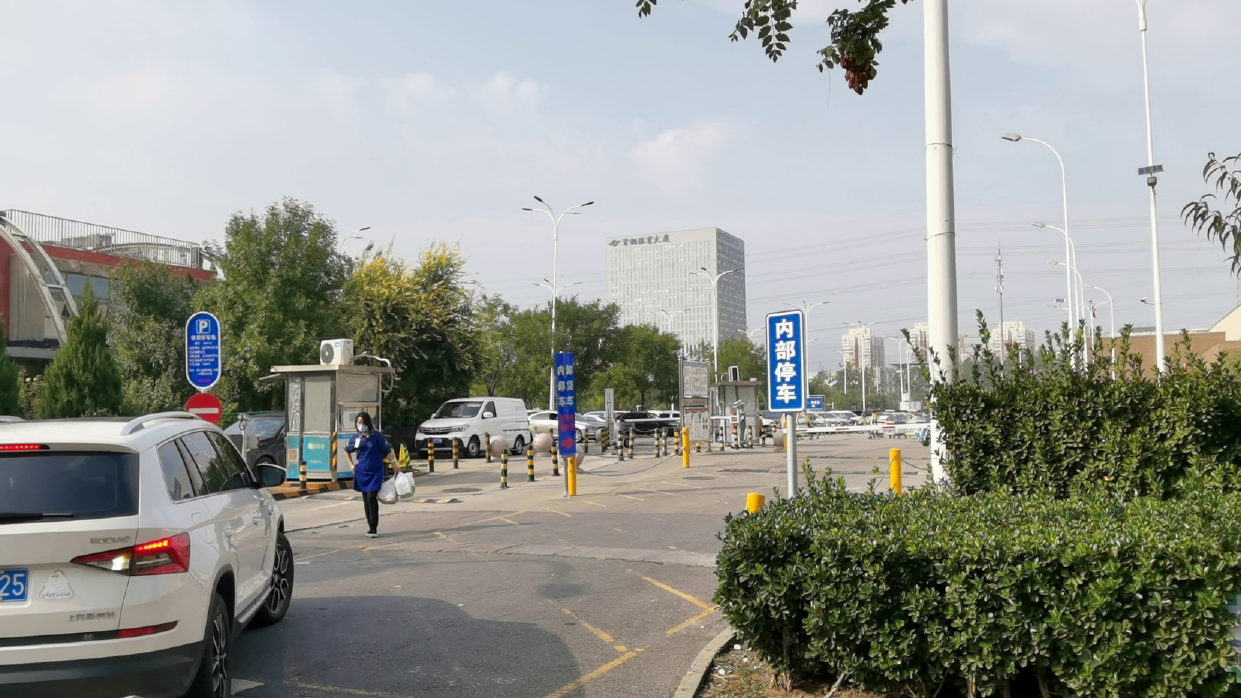} &
      \includegraphics[width=0.43\columnwidth]{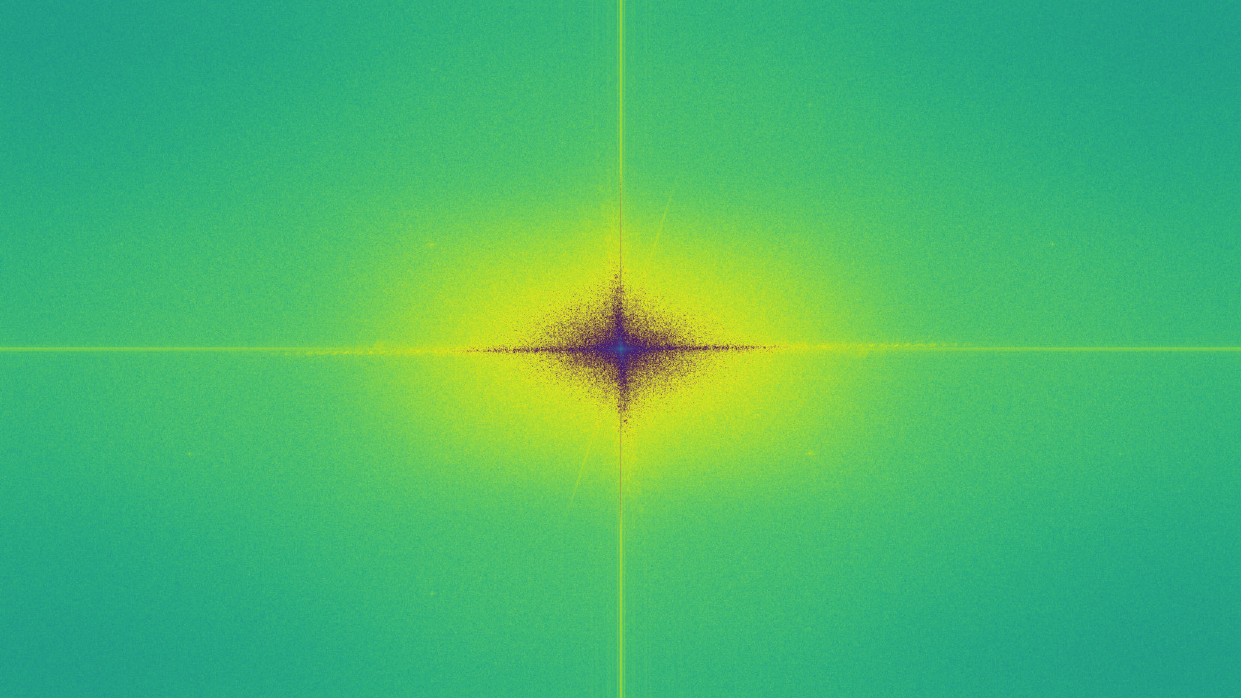} \\
      & & PSNR: 14.17 & PSNR: 21.44 \\
    \end{tabular}
  \end{minipage}
  \begin{minipage}{0.49\textwidth}
    \centering
    \begin{tabular}{cccc}
      \multirow{2}{*}{\rotatebox{90}{Deblur}} & 
      \raisebox{0.3\height}{\rotatebox{90}{Input}} &
      \includegraphics[width=0.43\columnwidth]{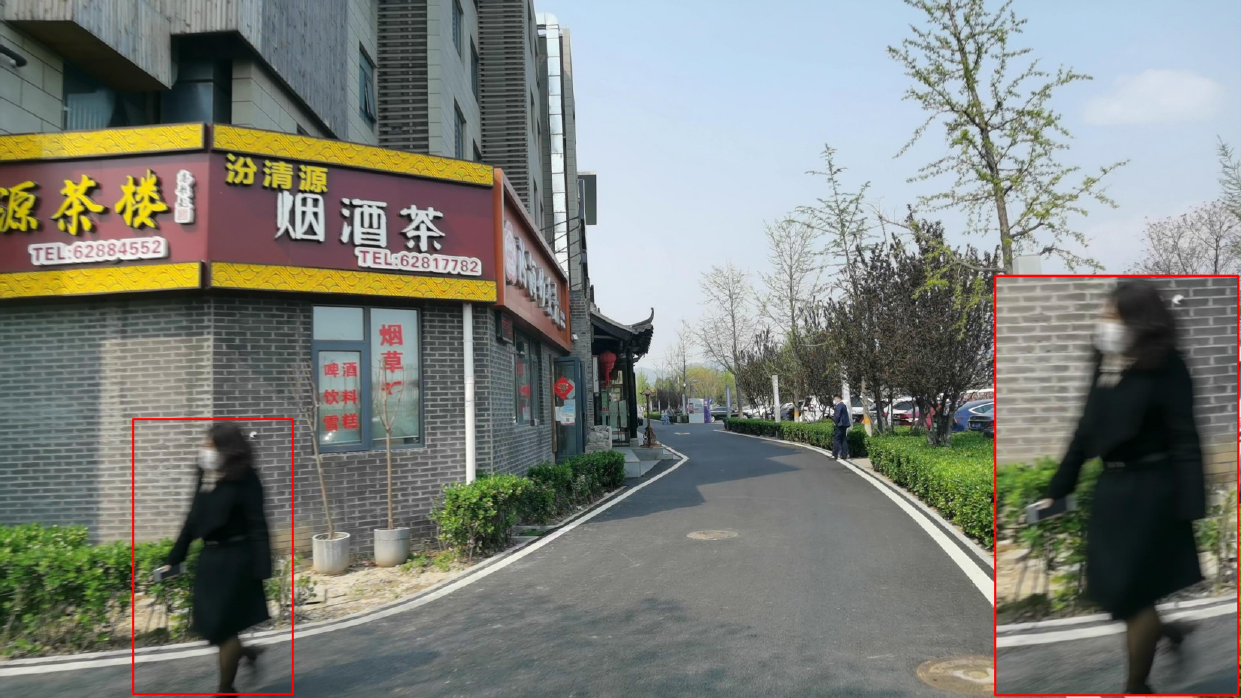} &
      \includegraphics[width=0.43\columnwidth]{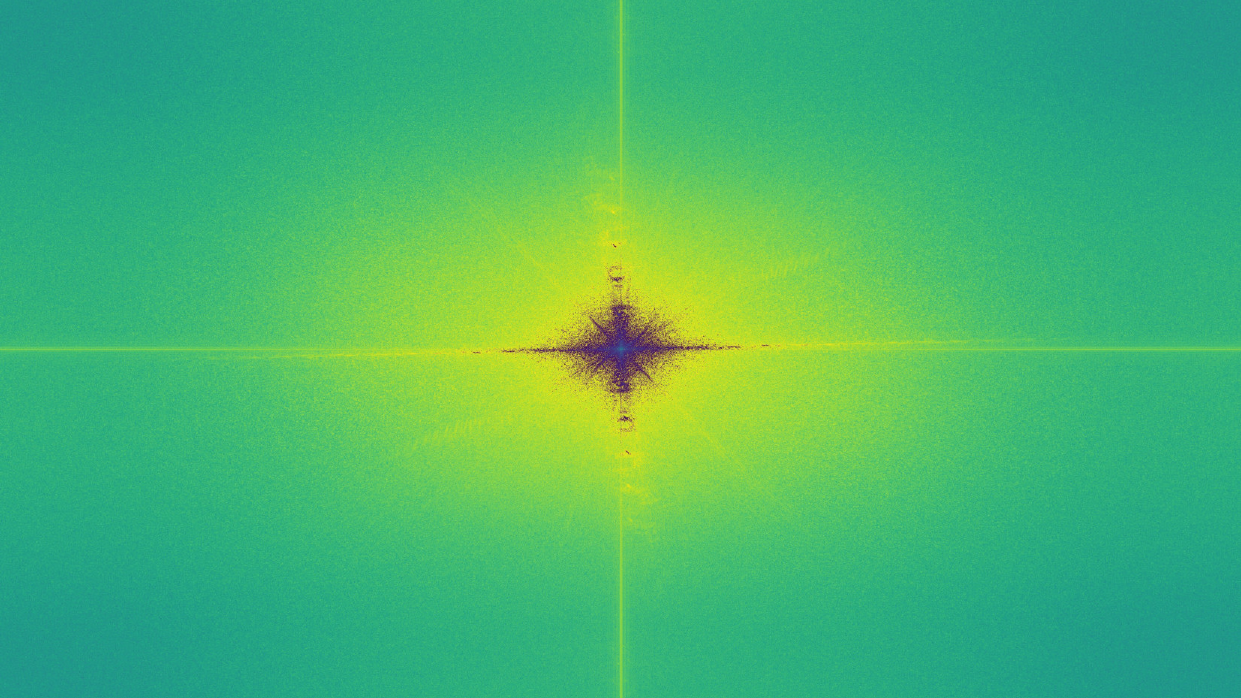} \\
      & \raisebox{\height}{\rotatebox{90}{GT}} & 
      \includegraphics[width=0.43\columnwidth]{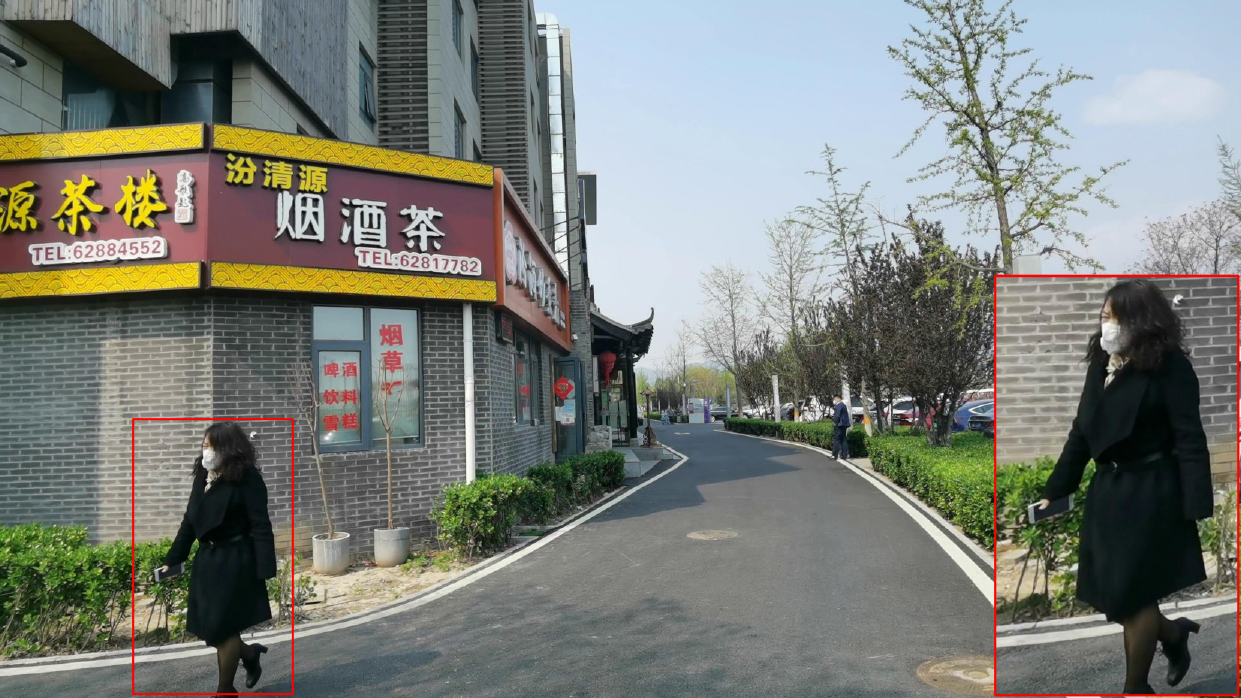} &
      \includegraphics[width=0.43\columnwidth]{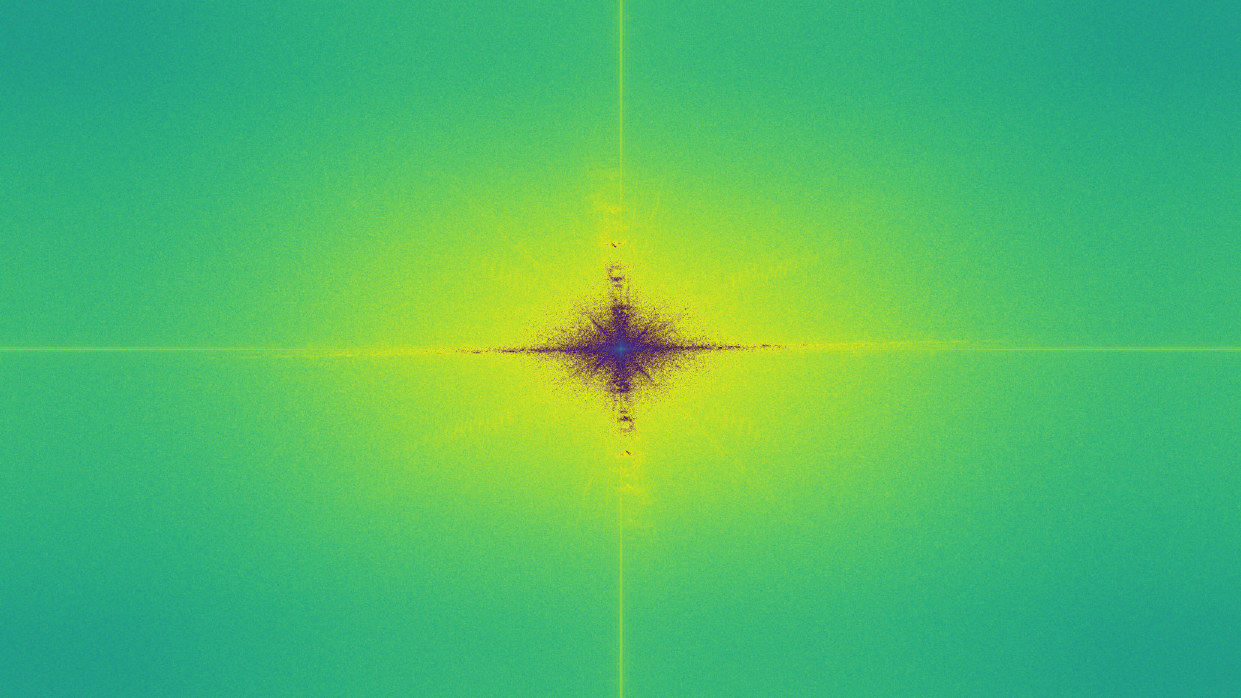} \\
      & & PSNR: 24.69 & PSNR: 18.15 \\
    \end{tabular}
  \end{minipage}
  
  \caption{Frequency analysis on the difference between distorted and clear images in the dehazing and deblurring task.}
  \label{fig:FourierAnalysis}
\end{figure}

To investigate this leverage, we conducted a frequency analysis on different tasks. We observed the difference between the clear image and the degraded image in both spatial and frequency domains, as shown in Fig. \ref{fig:FourierAnalysis}. For global haze, the frequency domain exhibits a higher PSNR compared to the spatial domain between the ground truth and input image, suggesting that the frequency domain more effectively captures the uniform degradation pattern. Conversely, for localized blur, the spatial domain shows a higher PSNR compared to the frequency domain, indicating that the spatial domain better reflects the localized degradation characteristics. This proves that in cases of local degradation, the spatial domain has a better representation of the relation between degraded and clear images.
 
In another aspect, Retinexformer \cite{cai2023retinexformer} proposed a retinex theory-based model to restore low-light images, and multiple researchers have followed the same concept \cite{10.1007/978-981-96-6596-9_30,LIU20251969}. Although these methods demonstrated state-of-the-art performance, they overlook the different nature of reflectance and illumination and do not explore the retinex theory as a general framework for image restoration beyond color correction.

Therefore, we propose \textbf{RetinexDual}, a novel approach that utilizes retinex theory as a generalized IR concept and addresses the duality nature of its decomposition (reflectance and illumination) separately using a two-branched architecture handling the complexity of UHD images. RetinexDual consists of two main networks: \textbf{S}cale \textbf{A}ttentive ma\textbf{MBA} (\textbf{SAMBA}) and \textbf{F}requency \textbf{I}llumination \textbf{A}daptor (\textbf{FIA}). SAMBA focuses on correcting the artifacts and distortions caused by degradation in the reflectance component by employing our tailored Scale-Adaptive Mamba block. Jointly, FIA alters the color distortion and exposure error in the illumination component by processing it in the Fourier domain. Comparative results in various tasks show state-of-the-art performance, with visual results supporting the same conclusion.
In addition, ablation studies prove the impact of different components in the approach.

Accordingly, the main contributions of our work are as follows:
\begin{itemize}
    \item We propose our novel \textbf{RetinexDual}, which to the best of our knowledge the first general UHD IR model that utilizes Retinex theory across multiple restoration tasks. It achieves this by decoupling an image into two components and processing each with a specialized sub-network tailored to their distinct characteristics.
    \item In the SAMBA branch, we design a \textbf{S}cale-\textbf{A}daptive \textbf{M}amba \textbf{B}lock (\textbf{SAMB}), that can adapt the selective scan process \textbf{using multiple scales of the features}, thus allowing non-causal modeling of mamba block to better rectify the distorted details of the reflectance component.
    \item In the FIA branch, we leverage the global context in the frequency domain to perform efficient color and exposure correction. Inside the FIA, we present the Fourier Correction Block (\textbf{FCB}), which has a reduced architectural complexity without compromising performance.
    \item \textbf{RetinexDual} outperforms existing methods in four UHD IR tasks qualitatively and quantitatively.
\end{itemize}

\section{Related Works}
In this section, we conduct a review of recent UHD IR methods and Retinex theory in image restoration.

\subsubsection{Ultra-High-Definition Image Restoration.} As Ultra-High-Definition imaging gained traction, efforts to enhance and recover UHD content started attracting significant focus \cite{wang2024uhdformer,Zhao_2025_CVPR,zou2024wavemamba,Wang_Zhang_Shen_Luo_Stenger_Lu_2023}. Wang et al. \cite{Wang_Zhang_Shen_Luo_Stenger_Lu_2023} proposed LLFormer that utilizes a fusion cross-attention block to seamlessly integrate information from different processing stages, while leveraging axis-based multi-head self-attention to process feature maps across multiple levels. Although, the model structure preserved the information in UHD images, it was computationally large to operate on common machinery. In Wave-Mamba \cite{zou2024wavemamba}, the authors suggested operating on high-frequency and low-frequency side by side using a different structure of mamba blocks to avoid losing information in downsampling. In ERR \cite{Zhao_2025_CVPR}, the authors built a 3-stage method for zero-frequency, low-frequency, and high-frequency information in the UHD image. Despite the ability of these methods to solve the computational problem and preserve the information, our frequency analysis concluded that in the case of localized degradation, as shown in Fig. \ref{fig:FourierAnalysis}, the spatial domain has a better representation of the distortion.

\subsubsection{Retinex Theory in Image Restoration.} Lately, LLIE research got redirected towards using retinex theory, through trying to conceptualize the learning-based methods to leverage having the noise mainly in the illumination component of the image \cite{cai2023retinexformer,10.1007/978-981-96-6596-9_30,LIU20251969}. In Retinexformer \cite{cai2023retinexformer}, the authors designed a retinex-based method that uses an illumination-guided transformer to achieve low-light enhancement. Following them, methods as RetinexMamba \cite{10.1007/978-981-96-6596-9_30}, Retinexformer+ \cite{LIU20251969}, ERetinex \cite{guo2025eretinex}, and MambaLLIE \cite{NEURIPS2024_30699996}, have been implementing the same approach with different deep learning methods. These methods have proven SOTA performance in low-light enhancement. However, the approaches were either operating on one of the retinex components of the image or used the same architectural structure for both components. Also, there was no investigation into the usage of retinex theory as a general IR concept. Therefore, understanding the difference between the retinex components and reconstructing the problem accordingly will enable it to achieve better performance.

\section{Methodology}
\begin{figure*}
\centering
\includegraphics[width=\textwidth]{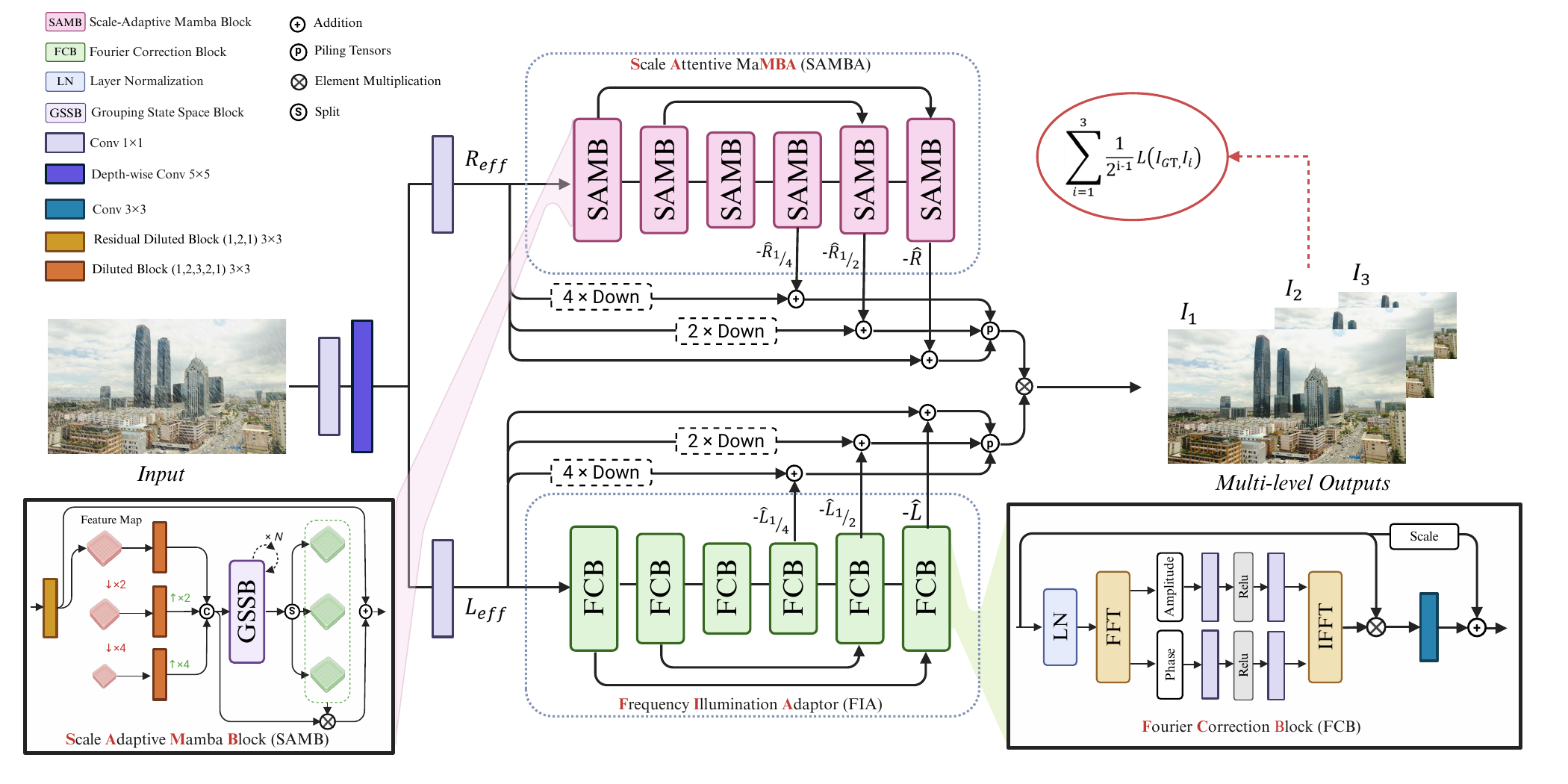} 
\caption{RetinexDual Overview. Based on retinex theory, it decomposes the UHD image into \(R_{eff}\) and \(L_{eff}\) and operates on them using 2 sub-networks: Scale Attentive MaMBA (SAMBA), and Frequency Illumination Adaptor (FIA), respectively. Noting that each output from a different level has a convolution layer before it for processing, which wasn't illustrated for simplicity.}
\label{fig:RD_overview}
\end{figure*}

In this section, we will deep dive into the RetinexDual approach that contains two sub-networks, SAMBA and FIA, as shown in Fig. \ref{fig:RD_overview}, and will illustrate how the training objective is reconstructed with the multi-level approach.
\subsection{RetinexDual Framework:}
As shown in Fig. \ref{fig:RD_overview}, RetinexDual has three main components, which are: Retinex decomposer, SAMBA, and FIA sub-networks. Starting with the problem formulation, Retinex theory simulates the human visual perception by the decomposition of the image \textbf{\( I \in \mathbb{R}^{H \times W \times 3} \)} to reflection component \(R \in \mathbb{R}^{H \times W \times 3}\) and illumination component \(L \in \mathbb{R}^{H \times W}\) as 
\begin{equation}
I = R \odot L
\end{equation}
where \(\odot\) denotes the element-wise multiplication. \(R\) represents the characteristics of the captured object, while \(L\) represents the light emerged on it. Taking into consideration the distortion of the image, it can be reformulated as
\begin{equation}
    \begin{aligned}
        I &= R_{eff} \odot L_{eff} \\
          &= (R + \hat{R}) \odot (L + \hat{L}),
    \end{aligned}
\end{equation}
where \(\hat{R}\) represents the artifacts or hidden features due to distortion, \(\hat{L}\) describes the color degradation or exposure error,and \(R_{eff}\), \(L_{eff}\) represent the image with the corruption. Given a distorted image \(I \in \mathbb{R}^{H \times W \times 3}\), the retinex decomposer employs two convolution layers to extract deeper features. Then a \(1 \times 1\) convolution layer is used to construct \(R_{eff}\) and \(L_{eff}\) separately to aggregate them from the features of the input image \(I\). As we assume that our retinex decomposer \(\psi_d\) will derive \(R_{eff}\) and \(L_{eff}\), we can reconstruct our mathematical model for general image restoration as follows:
\begin{equation}
    \begin{aligned}
    R_{eff},L_{eff}&=\psi_d (x),\kern0.1em  \\R= R_{eff} -\hat R ,&\kern0.1em L =  L_{eff}-\hat L \\
    \hat{R} = -\mathcal{S}(R_{eff}),&\kern0.4em \hat{L} = -\mathcal{F}(L_{eff}), \\ I = R& \odot L,
    \end{aligned}
\end{equation}
Where \(\mathcal{S}\) , \(\mathcal{F}\) are SAMBA and FIA sub-networks respectively, and \(I\) is the corruption-free image.

Multiple researchers have attempted to adapt their framework to process \(I \odot \bar L\) \cite{cai2023retinexformer,10.1007/978-981-96-6596-9_30,LIU20251969}. However, this procedure was ideal for low-light enhancement, as the type of restoration is well-defined. In our approach, we aim to generalize it further by operating on the main components of the image \(R_{eff}, L_{eff}\) directly, which gives our framework the separation needed for adaptation and correction of distortion. Although some other approaches, such as Retinex-Diff \cite{yi2023diff}, and RAUNA \cite{10174279}, had two sub-networks with the same architecture to alter \(L\),\(R\) components separately, they did not build the sub-networks based on the nature of the components themselves. Alternatively, our model is designed with two complementary sub-networks, which constitute our first contribution.

\subsection{Scale Attentive MaMBA (SAMBA):}
Multiple attempts have been made to adapt Mamba for vision tasks, specifically IR tasks \cite{zou2024wavemamba,Li_2025_CVPR}. Most of these attempts focused on modifying the scanning strategy on the image to achieve better performance by multi-directional scans and different scanning schemes. However, they overlooked the inductive bias of spatial tasks, as the relation between pixels isn't sequential and extends to the global structure of the image. In causal modelling of Mamba, the $i$-th pixel is limited to observing only the $i-1$ pixels of the whole image, unable to leverage similar pixels globally. Moreover, a strong correlation with significant redundancy is indicated by the similarity of various scanned sequences on all testing datasets, reaching even over 0.7 \cite{Guo_2025_CVPR}. Additionally, according to Mamba \cite{gu2024mambalineartimesequencemodeling}, its causal nature tends to cause long-range decay defects.

Accordingly, we designed SAMBA, an architecture that relies on a coarse-to-fine mechanism, enabling the Mamba approach to adapt the scan strategy based on the relation between different regions of the input features on different scales. Our sub-network is an encoder-decoder-based model with 3 Scale Adaptive Mamba Blocks (SAMB), each as shown in Fig. \ref{fig:RD_overview}. 

\subsubsection{Scale Adaptive Mamba Block (SAMB).} Our SAMB starts with a  Residual Diluted Block (RDB) for feature refining and extraction. Then, our second contribution lies in using three multi-scale levels of the feature map \((1,  1/2, 1/4)\), which are further processed by diluted convolution, concatenated together, and passed to the Group State Space Block (GSSB) eventually. Having different scales of feature maps helps in capturing the relationship between features far from each other in the image, overcoming the drawback of convolution (the inductive bias of spatial locality).

SAMB Block can be expressed as:
\begin{equation}
    \begin{aligned}
    x_{s} =& RDB(x_{in})\\
    x_{0}, x_{1}, x_{2} = &Up(DB(x_{s},x_{s \downarrow},x_{s\downarrow\downarrow})))\\
    \hat{x}_{0}, \hat{x}_{1}, \hat{x}_{2} =& \mathcal{S}(GSSB(\mathcal{C}[x_{0}, x_{1}, x_{2}]))\\
    x' = (&\sum_{i=0}^{2} x_{i} \odot \hat{x}_{i})+x_{s},
    \end{aligned}
\end{equation}
where \(RDB\), \(DB\) refer to the residual diluted block and the diluted convolution block, respectively. \(Up\) is the bilinear interpolation of the output to the original dimensions. \(S\), \(C\) refer to split operation and channel concatenation.

\begin{figure}
\centering
\includegraphics[width=\columnwidth]{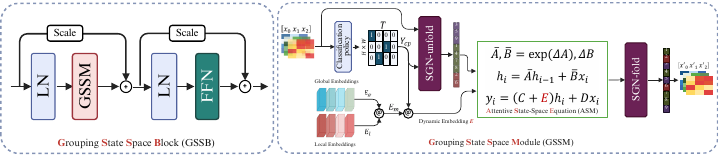} 
\caption{The architecture of Grouping State Space Block (GSSB)}
\label{fig:GSSM_overview}
\end{figure}
\subsubsection{Group State Space Block (GSSB).} As shown in Fig. \ref{fig:GSSM_overview}, We utilize Mamba's ability to analyze long-distance sequences linearly and its lower computing complexity compared to Transformers. We follow the efficient token mixer of the transformer that has been adopted by many researchers \cite{10.1007/978-981-96-0911-6_10}, which can be expressed mathematically as follows:
\begin{equation}
    \begin{aligned}
    \hat x &= GSSM(LN(x)) + s \cdot x, \\
    x' &= FFN(LN(\hat x)) + s' \cdot \hat x,
    \end{aligned}
\end{equation}
where \(LN\), \(FFN\) refer to layer normalization, convolutional feed forward layer. \(s, s' \in \mathbb{R}^{1 \times 1 \times (3C)}\) are learnable weights to scale the factor of residual.

\subsubsection{Group State Space Module (GSSM).} Inspired by MambaIRv2 \cite{Guo_2025_CVPR}, we build our Group State Space Module (GSSM) to solve the causal modeling problem in Mamba. Given \(x \in \mathbb{R}^{H \times W \times (3C) }\), we start by using positional encoding \cite{DBLP:journals/corr/abs-2102-10882} on \(x\) to preserve the spatial information structure. Aiming to overcome the causal nature of Mamba, we introduce local \(E_l\) and global \(E_g\) embeddings in the block and across the model, respectively, to represent pixels with similar semantics into \(C\) component of the state space model as image-specific embedding \(E = (E_lE_g)Y_{cp}\). \(Y_{cp}\) is a classification policy, computed to represent the probability of each embedding being sampled by flattened input \(x'_i, i= 1,2 \dots L\). This will provide the missing information due to unseen regions for the model.
Going deeper into the network, we apply Semantic Guided Neighboring (SGN) proposed in \cite{Guo_2025_CVPR} as a scan strategy that arranges the sequence according to the semantic similarity of different regions. Accordingly, SGN unfolds them into a semantic sequence for Attentive State-space Equation (ASE) as shown in Fig. \ref{fig:GSSM_overview}. Noting that in our case, this semantic similarity will go beyond the spatial locality and capture the relations on different scales of the image. 
Eventually, the dynamic embedding \(E\) is then added to the \(C\) in the Attentive State-space Equation as follows:
\begin{equation}
    \begin{aligned}
        h'(t) &= \bar{A}h(t) + \bar{B}x(t), \\
        y(t) &= (C+E)h(t) + Dx(t),
    \end{aligned}
\end{equation}
where \(E\) is the image-specific embedding. 
Finally, the SGN folds the output to its original structure, and the feature map is reconstructed to \(x' \in \mathbb{R}^{H \times W \times (3C) }\).

\subsection{Frequency Illumination Adaptor (FIA):}
Converting UHD images from the spatial domain to the frequency domain has been proven to be effective with illumination correction and color distortion restoration \cite{Li2023ICLR,10.1007/978-981-96-0911-6_10,lv2024fourier}. Specifically, due to it ability to represent the global context in high and low frequencies, the Fourier transform is more suited for retrieving texture detail information. Accordingly, we propose a Frequency Illumination Adaptor (FIA) that works mainly in the Fourier domain to alter the illumination component of the retinex decomposition.

\subsubsection{Fourier Correction Block (FCB).} As observed in Fig. \ref{fig:RD_overview}, given input feature \(x \in \mathbb{R}^{H \times W \times C}\), we first pass it through layer normalization, then use FFT to decompose \(x\) into amplitude \(A\) and phase \(P\). Each component is processed using two \(1 \times 1 \) convolutions separated by a ReLU activation function. The result of the two branches reassembles the image to the spatial domain using IFFT, forming \(\hat x\). To avoid instability and apply adaptive scaling, we multiply \(\hat x\) with the input and use a final \(3 \times 3 \) convolution layer followed by a scaled skip connection. Maintaining such a compact design that includes only 0.2M parameters serves as our third contribution. This network can be represented as :
\begin{equation}
    \begin{aligned}
        \hat x = IFFT(Co&nv(RELU(Conv(FFT(LN(x)))))), \\
    x' &= Conv3(x \odot \hat x)+ s \cdot x,
    \end{aligned}
\end{equation}
where \(Conv\), \(Conv3\) are \(1 \times 1\) and \(3 \times 3\) convolution layers, \(s\) are learnable weights as scale factor for skip connection .

\subsection{Multi-level Training Objective:}
Aiming for generalization over all UHD IR tasks, We utilize four different training objectives to enhance our network's performance: Charbonnier loss \(\mathcal{L}_{cb}\)  \cite{DBLP:journals/corr/abs-1710-01992}, FFT loss \(\mathcal{L}_{FFT}\), SSIM loss \(\mathcal{L}_{ssim}\) \cite{1284395}, and perceptual loss \(\mathcal{L}_{p}\) \cite{8578166} using VGG16 network \cite{simonyan2015deepconvolutionalnetworkslargescale}. Given \(I, \hat I\) refer to the ground truth and predicted image, respectively, the training objective is defined as follows:

\begin{equation}
    \begin{aligned}
        \mathcal{L} = \lambda_{cb}\mathcal{L}_{cb}(I,\hat I)&+\lambda_{FFT}\mathcal{L}_{1}(FFT(I),FFT(\hat I)) \\
        +\lambda_{ssim}&\mathcal{L}_{ssim}(I,\hat I)+\lambda_{p}\mathcal{L}_{p}(I,\hat I),
    \end{aligned}
\end{equation}
where \(\lambda_{cb},\lambda_{FFT},\lambda_{ssim},\lambda_{p}\) are weight factors for each of the losses with 1, 0.1, 0.5, and 0.4, respectively.
Aiming to inject the optimization into the low levels of the model, we follow the deep supervision strategy by having output from multiple levels of the framework with different scales \(I_1,I_2,I_3\) as shown in Fig. \ref{fig:RD_overview}, while making sure that each scale is reconstructed following the retinex theory to avoid violating its concept. Furthermore, we applied a scaling factor on each level to make sure that its contribution to the optimization is controlled. This strategy can be expressed as:
\begin{equation}
    \begin{aligned}
        \mathcal{L}_{multi-level} = \sum_{i=1}^{3} \frac{1}{2^{i-1}} \mathcal{L}(I_i, \hat I_i),
    \end{aligned}
\end{equation}
where \(I_i\) represents the ground truth in different scales using bilinear interpolation.

\section{Experiments}

In this section, we present the benchmarks used in our study, the implementation settings, and the evaluation metrics employed to compare our method with state-of-the-art approaches, both qualitatively and quantitatively. In addition, we report the results of our ablation studies. Noting that more qualitative and quantitative comparative results are presented in the supplementary materials.
\subsection{Experimental settings:}
\subsubsection{Datasets.}
RetinexDual is evaluated using four UHD datasets representing different image restoration tasks. We assess the low-light enhancement task on the \textbf{UHD-LL} dataset \cite{Li2023ICLR}. Image dehazing and Image deblurring tasks are evaluated using the \textbf{UHD-Haze} and the \textbf{UHD-Blur} datasets \cite{wang2024uhdformer}, respectively. We evaluate Image deraining task using the \textbf{4K-Rain13K} dataset \cite{chen2024towards}.
\subsubsection{Implementation details.}
We train RetinexDual using four NVIDIA H100 GPUs with the following settings. Using AdamW optimizer \cite{loshchilov2019decoupledweightdecayregularization} with an initial learning rate of $1e^{-4}$ which is gradually reduced to $1e^{-7}$ through cosine annealing learning rate scheduler. The patch size used is 768×768 with batch size of 6.
\subsubsection{Evaluation.}
The evaluation of different architectures used on the four datasets is done mainly using PSNR and SSIM \cite{1284395}. The number of parameters is also reported to reflect the complexity of different architectures. 

\begin{figure*}
  \centering
  \setlength{\tabcolsep}{2pt}
  \begin{tabular}{c@{\hspace{0.5mm}}c@{\hspace{0.5mm}}c@{\hspace{0.5mm}}c@{\hspace{0.5mm}}c@{\hspace{0.5mm}}c@{\hspace{0.5mm}}c}
    \includegraphics[width=0.135\textwidth]{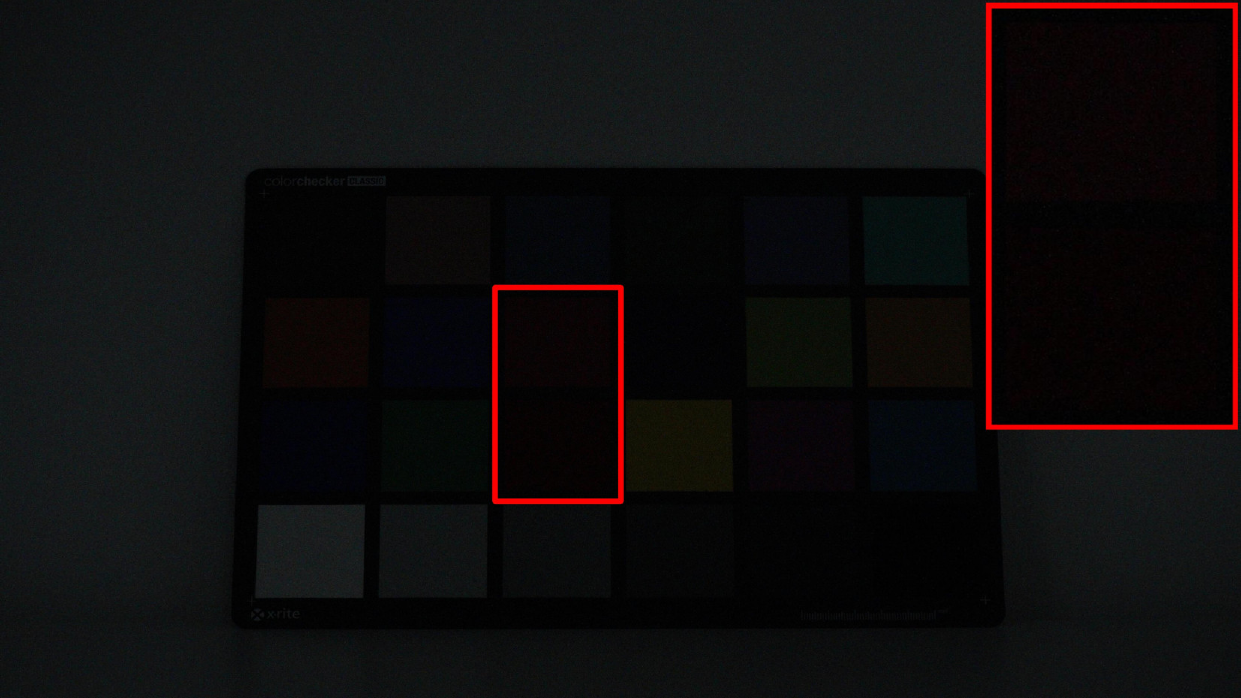} &
    \includegraphics[width=0.135\textwidth]{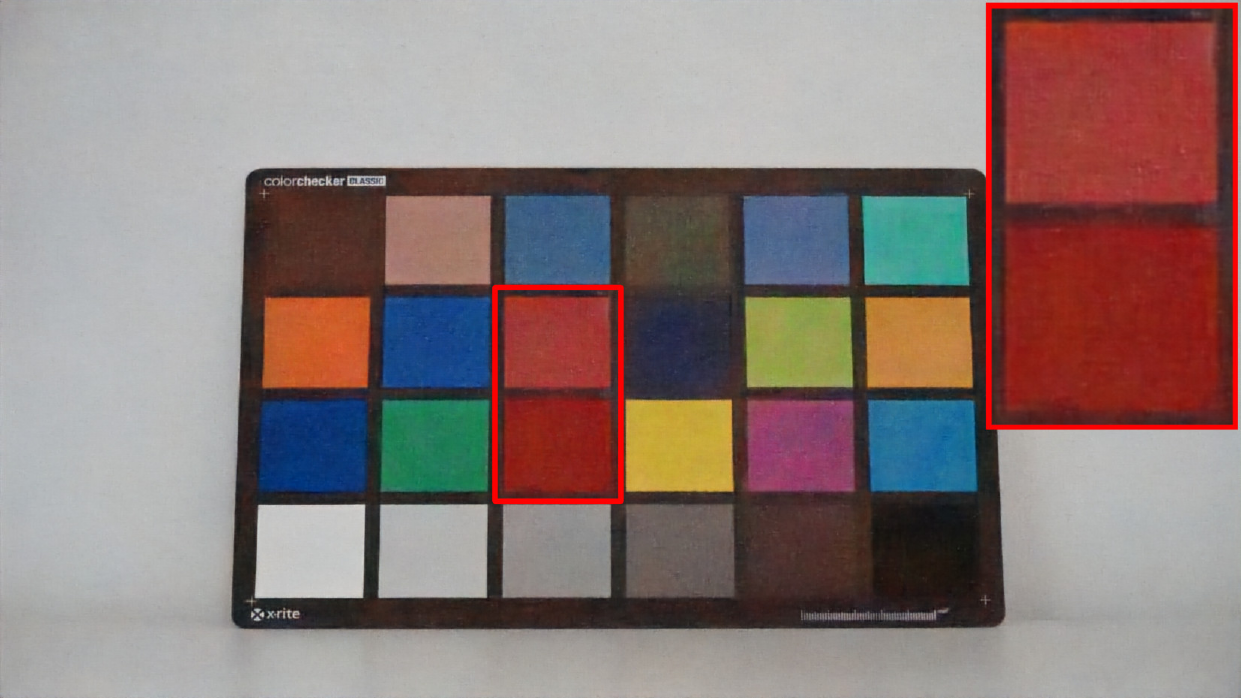} &
    \includegraphics[width=0.135\textwidth]{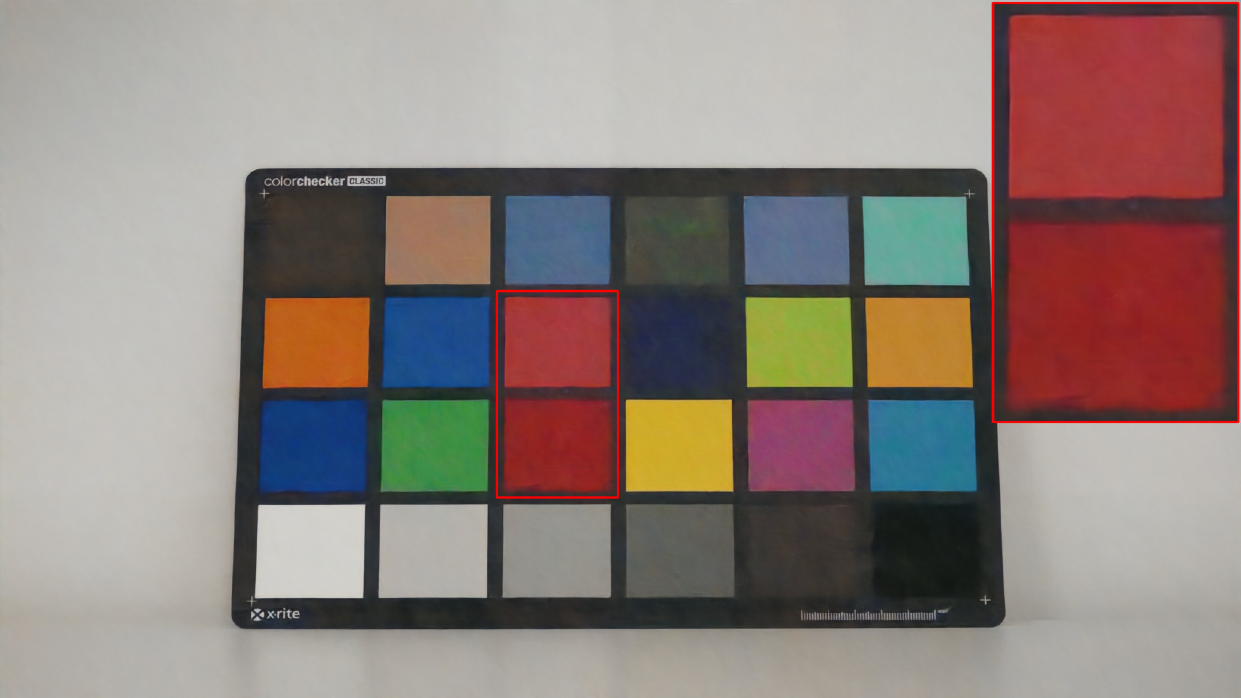} &
    \includegraphics[width=0.135\textwidth]{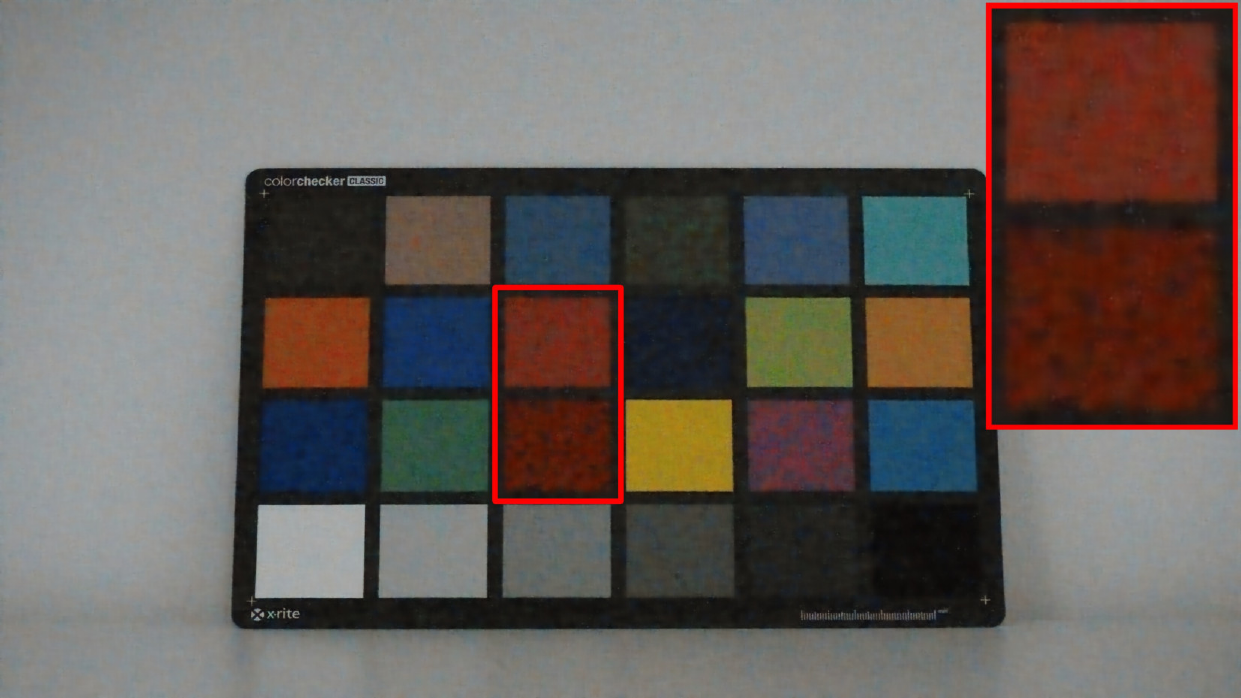} &
    \includegraphics[width=0.135\textwidth]{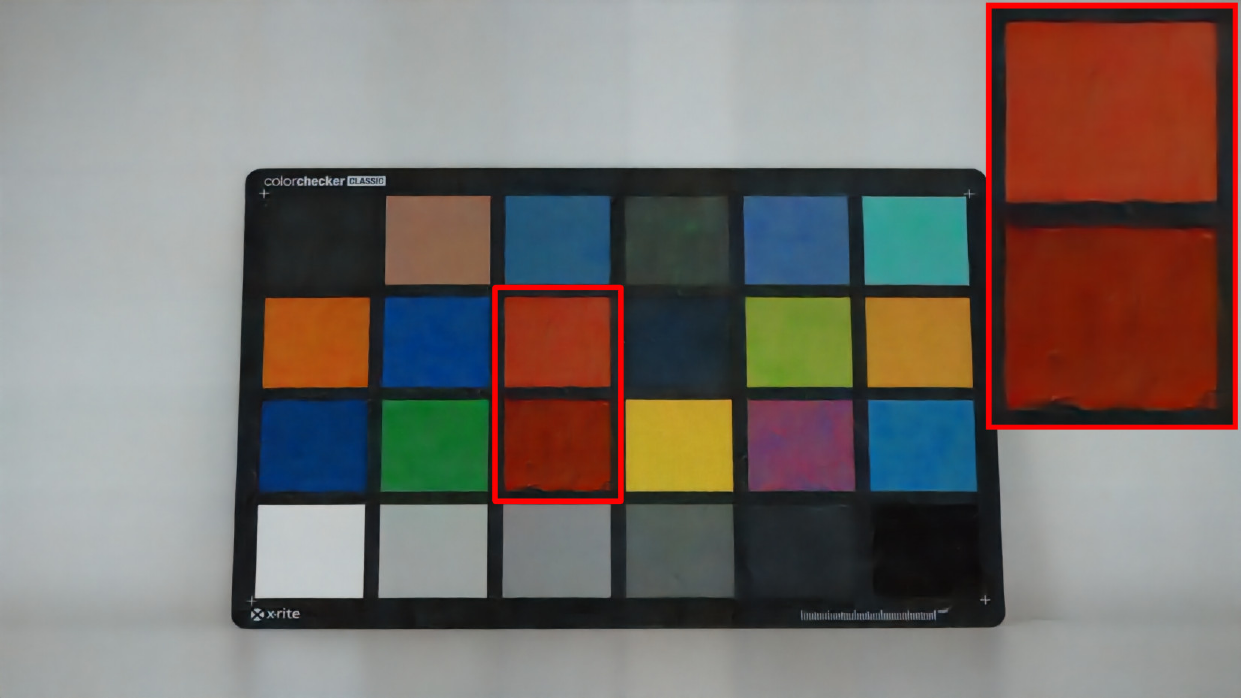} &
    \includegraphics[width=0.135\textwidth]{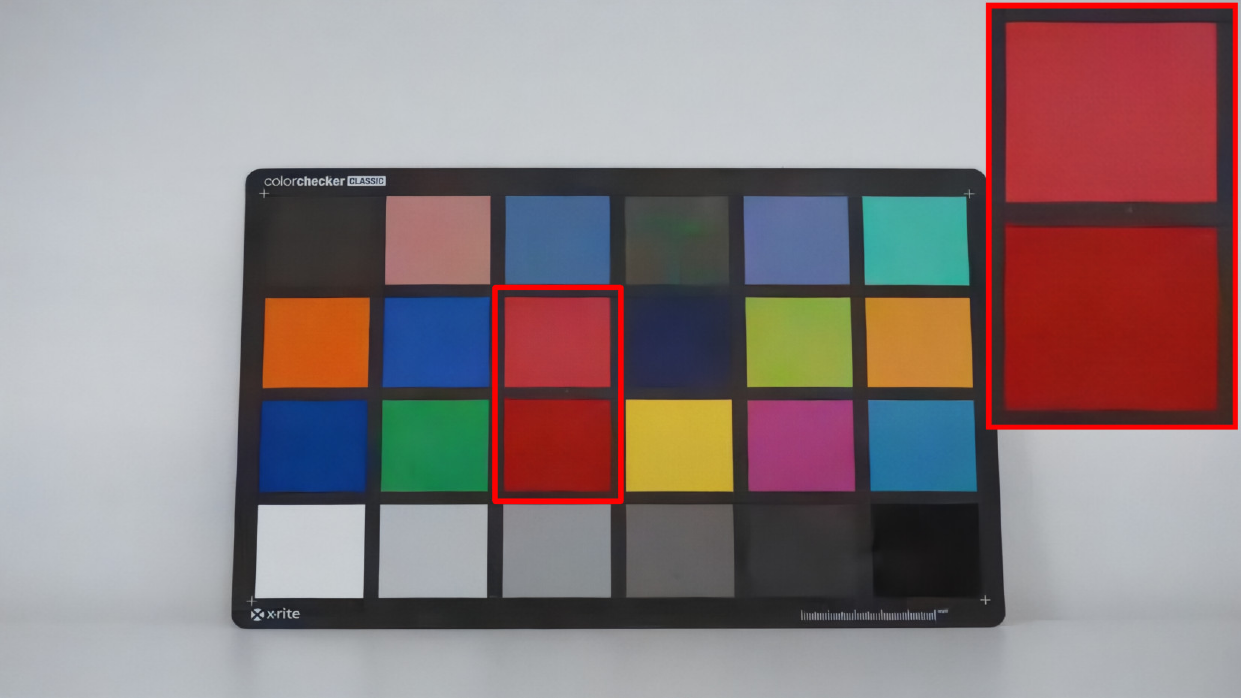} &
    \includegraphics[width=0.135\textwidth]{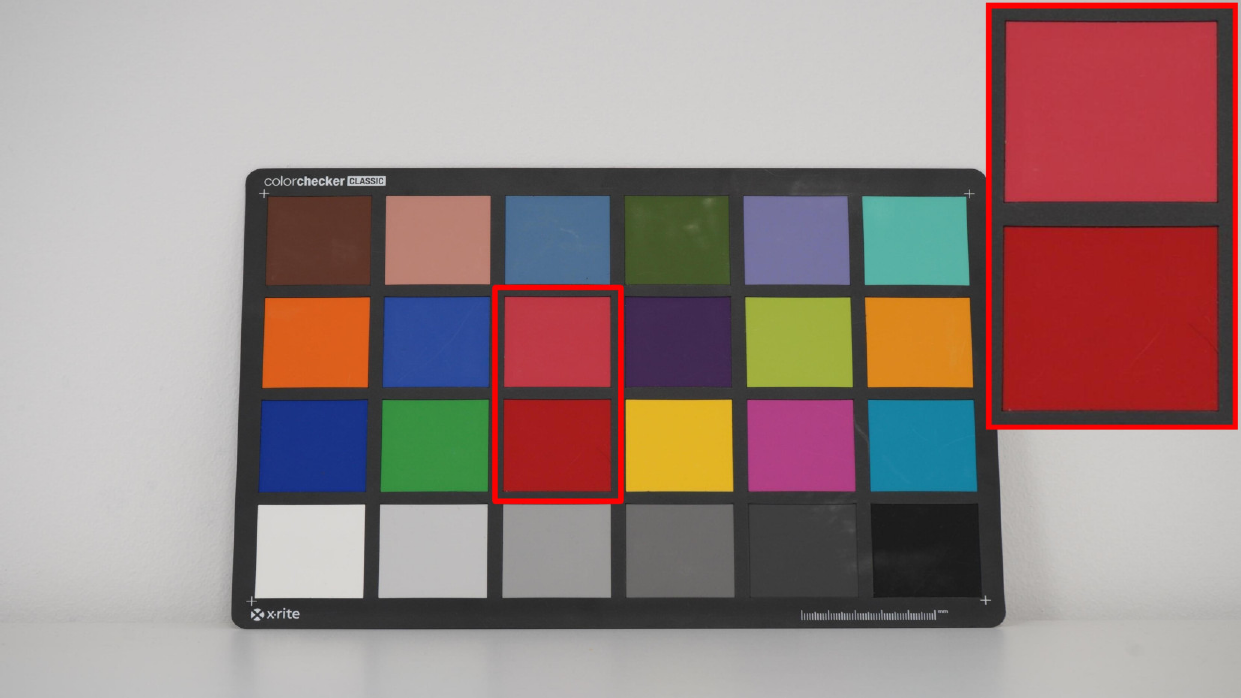} \\
    Input & UHDFour & WaveMamba & UHDForm & ERR & Ours & GT \\

    \includegraphics[width=0.135\textwidth]{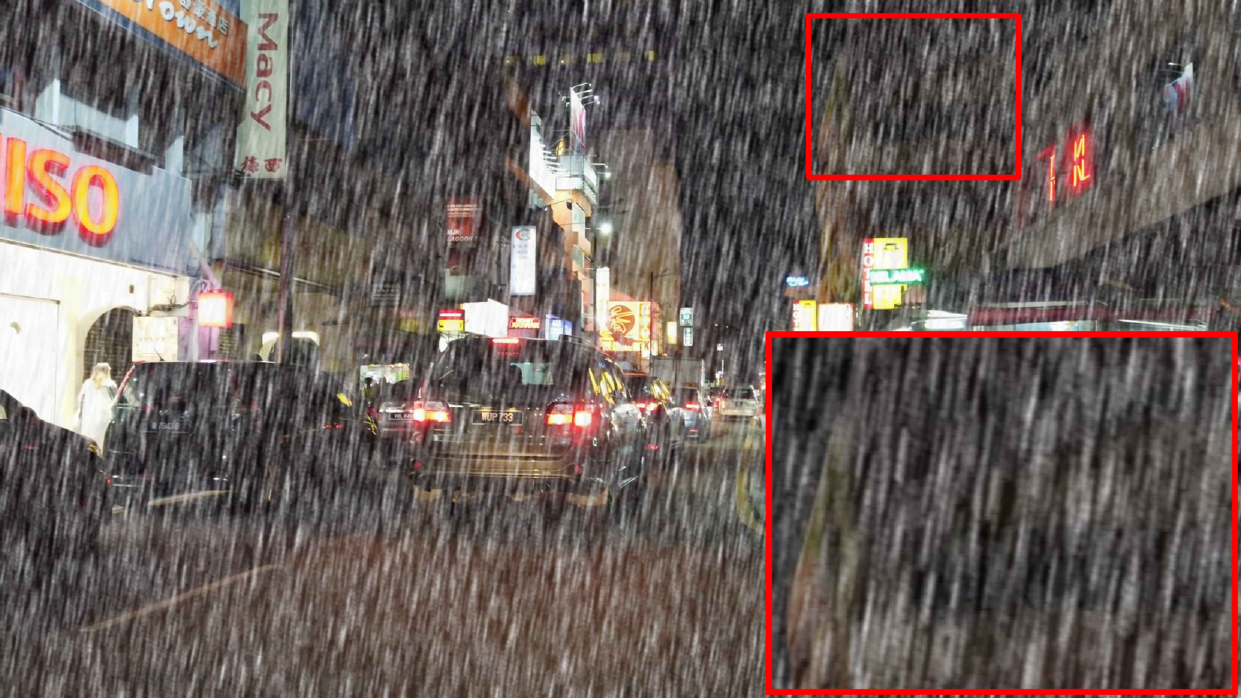} &
    \includegraphics[width=0.135\textwidth]{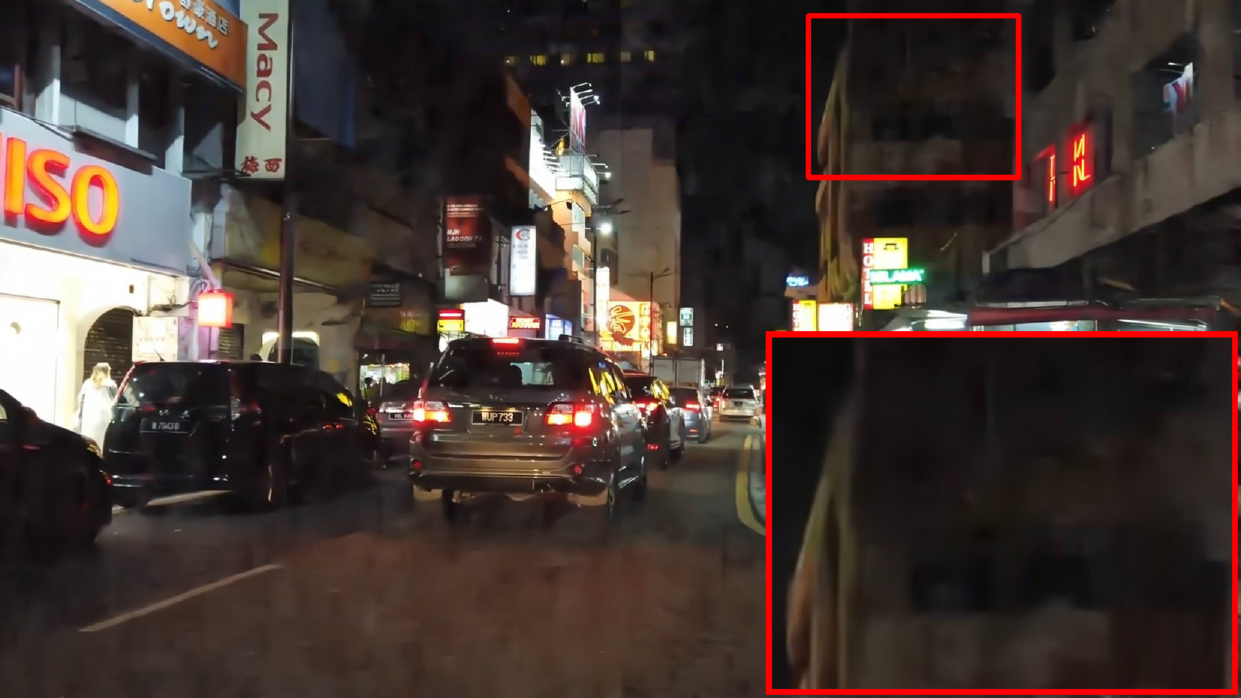} &
    \includegraphics[width=0.135\textwidth]{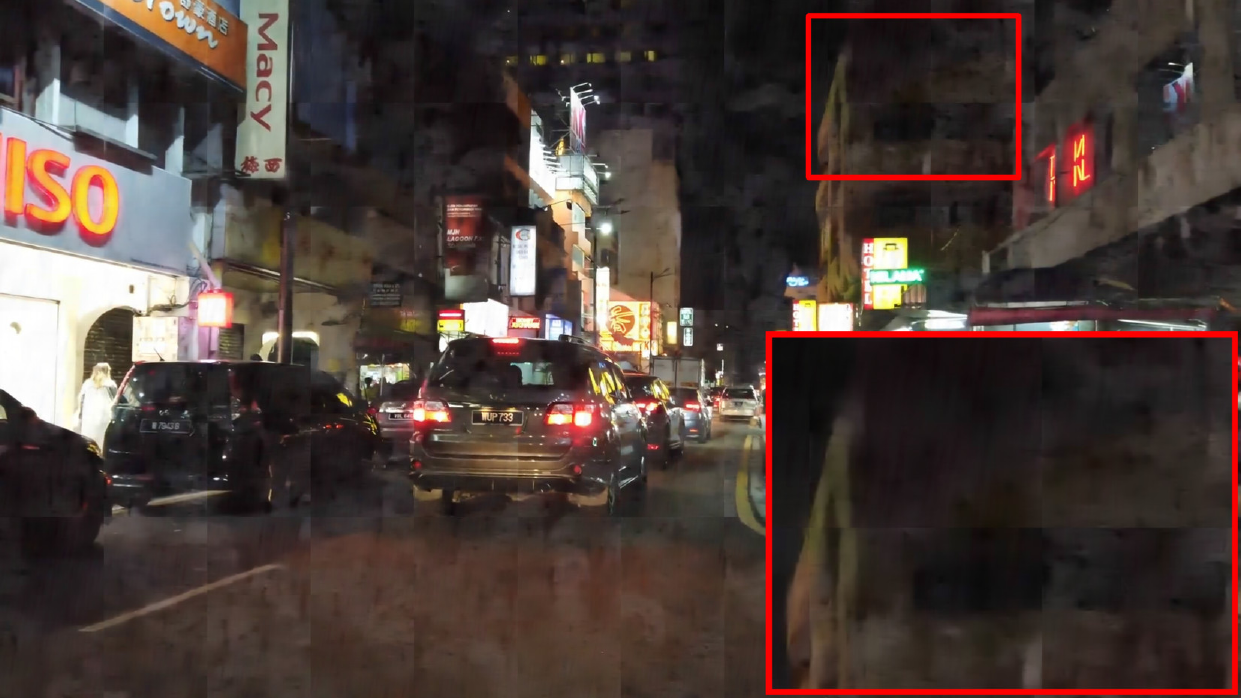} &
    \includegraphics[width=0.135\textwidth]{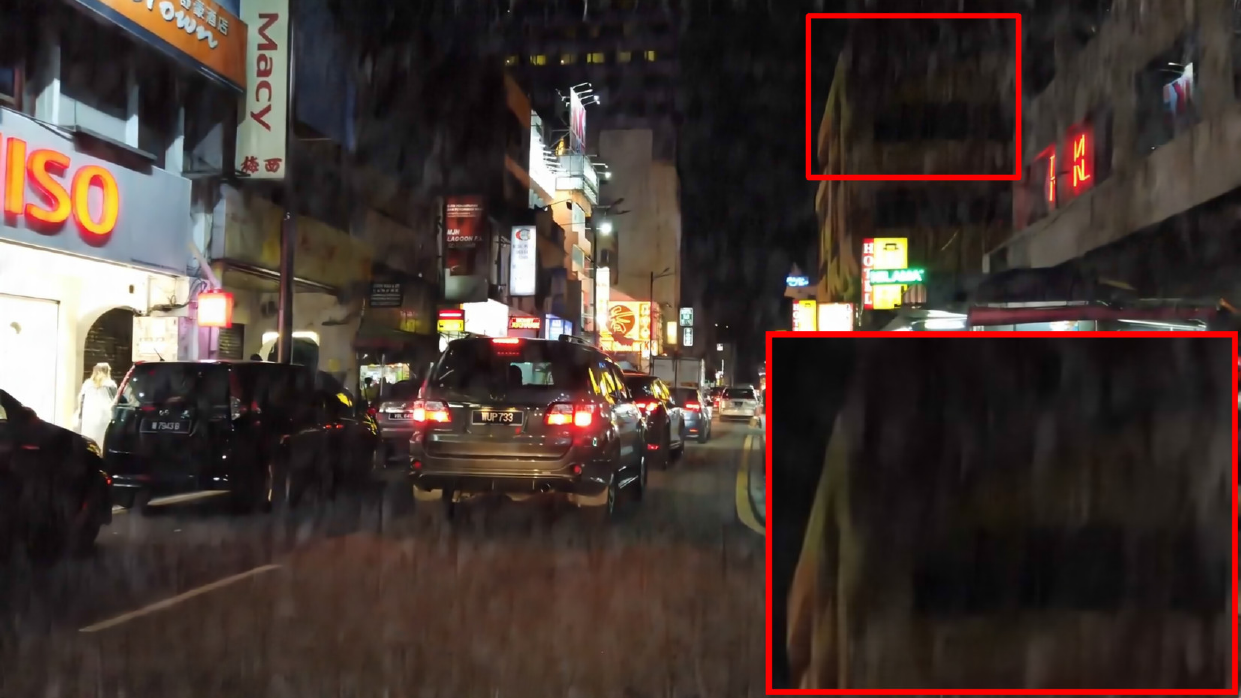} &
    \includegraphics[width=0.135\textwidth]{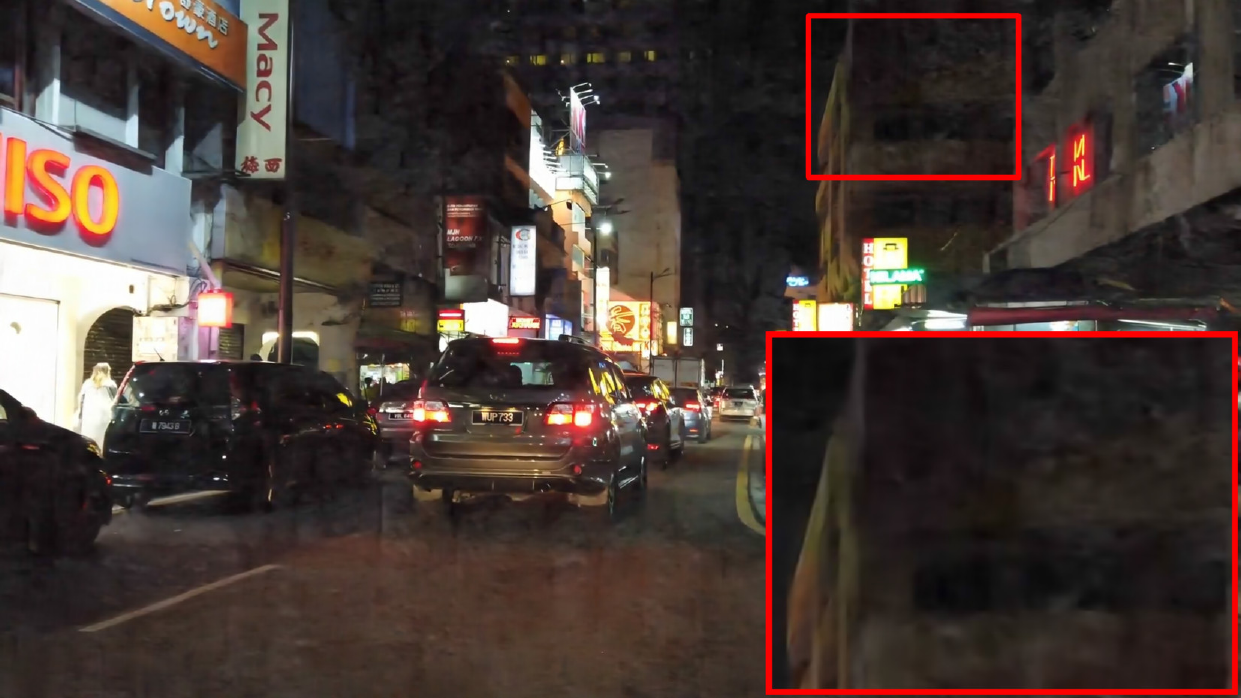} &
    \includegraphics[width=0.135\textwidth]{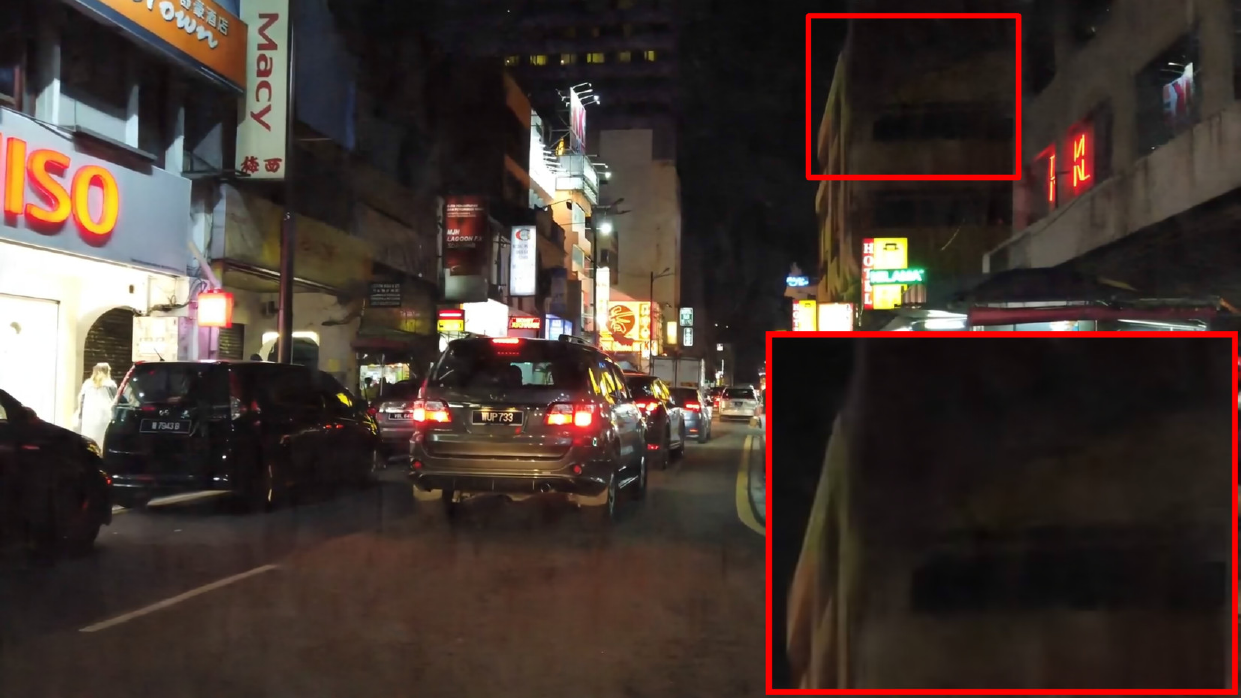} &
    \includegraphics[width=0.135\textwidth]{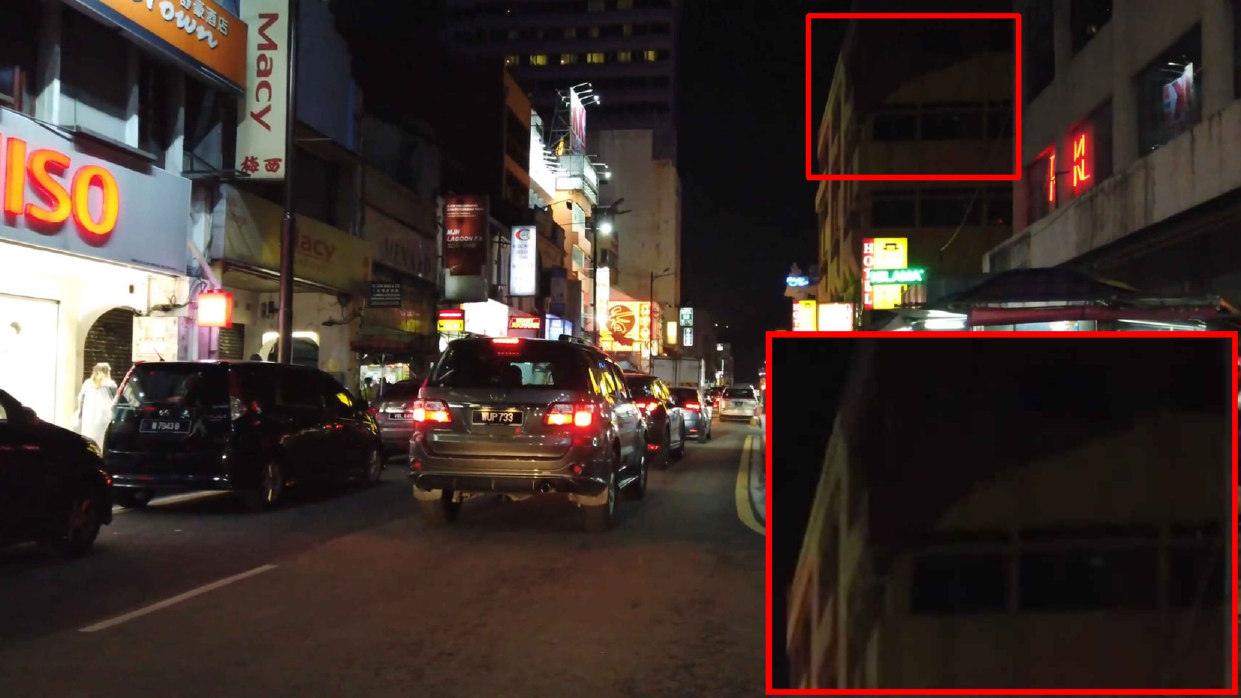} \\
    Input & DRSform & UDRS2Form & UDR-Mixer & ERR & Ours & GT \\

  \end{tabular}

  \caption{Visual comparison between different architectures on UHD-LL and 4K-Rain 13K. The first row shows results from UHD-LL \cite{Li2023ICLR} test (1188) while the second row shows results from 4K-Rain13K dataset \cite{chen2024towards} test (66).}
  \label{fig:VisualComparison1}
\end{figure*}

\begin{figure*}
  \centering
  \setlength{\tabcolsep}{1pt}
  \begin{tabular}{cccccc}
    \includegraphics[width=0.16\textwidth]{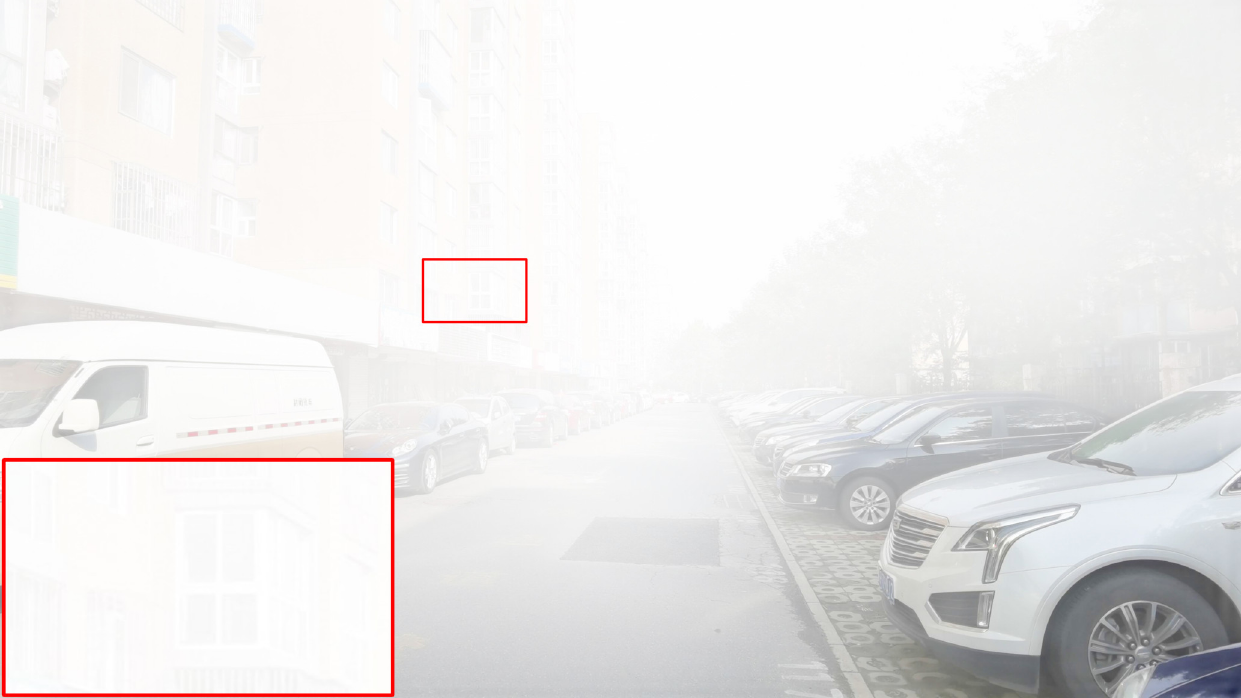} &
    \includegraphics[width=0.16\textwidth]{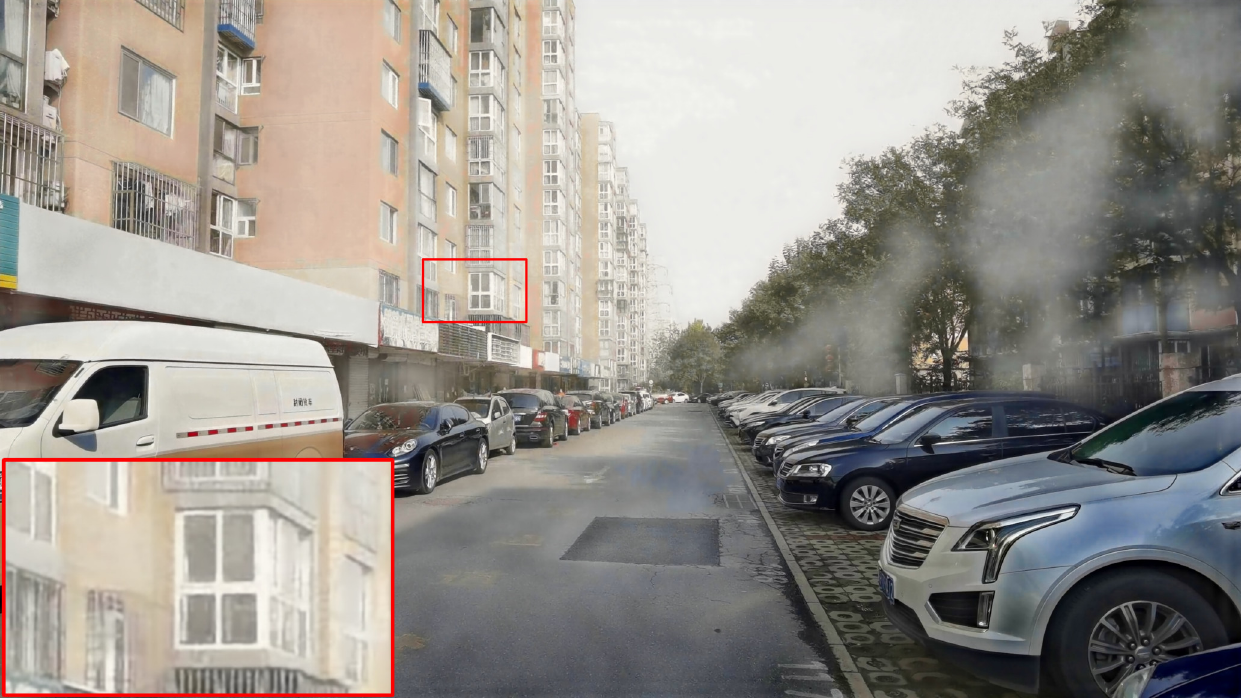} &
    \includegraphics[width=0.16\textwidth]{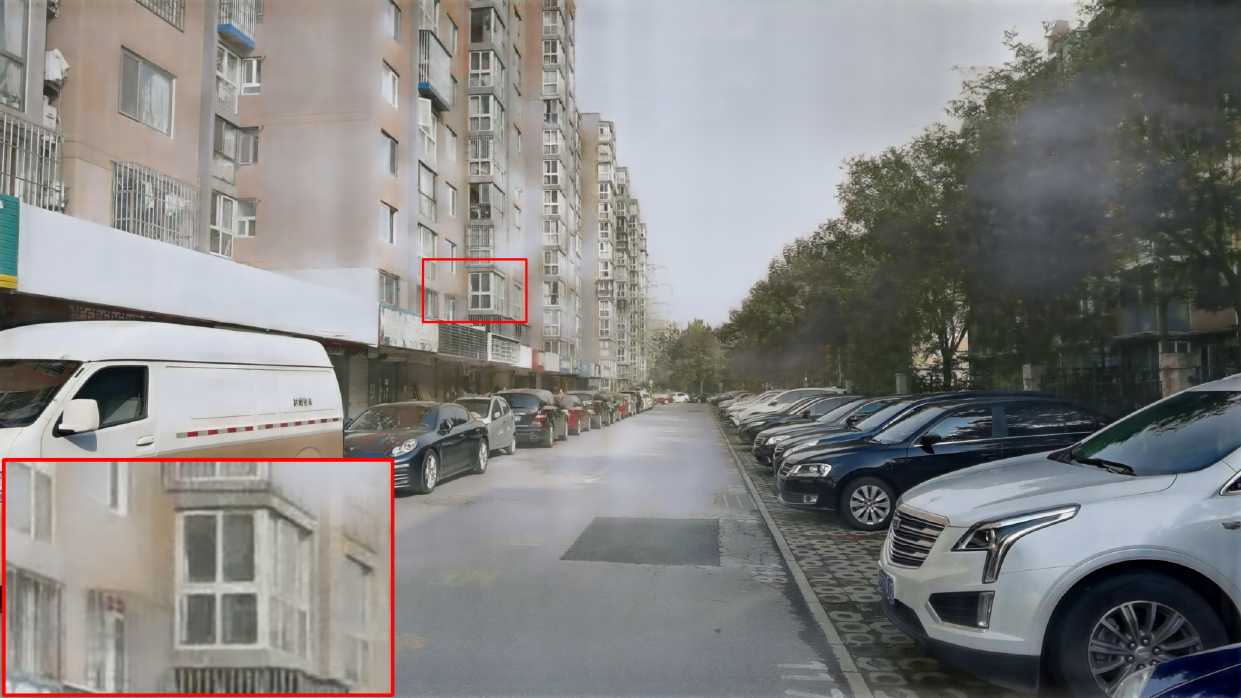} &
    \includegraphics[width=0.16\textwidth]{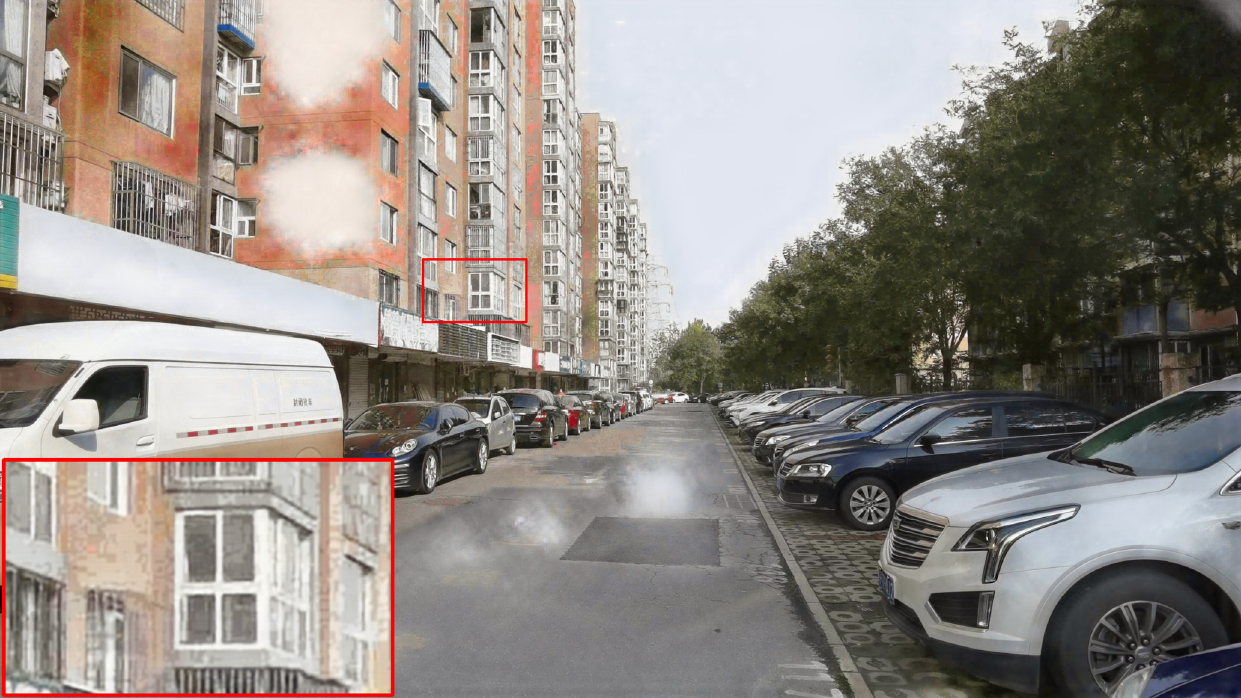} &
    \includegraphics[width=0.16\textwidth]{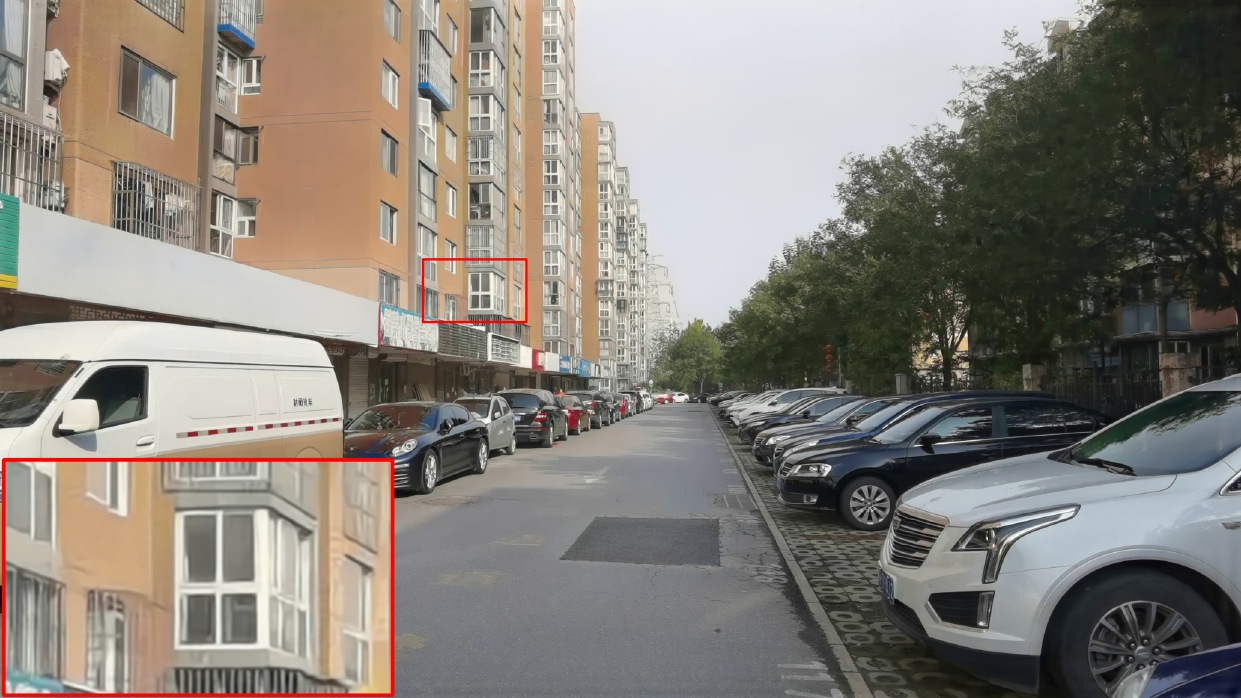} &
    \includegraphics[width=0.16\textwidth]{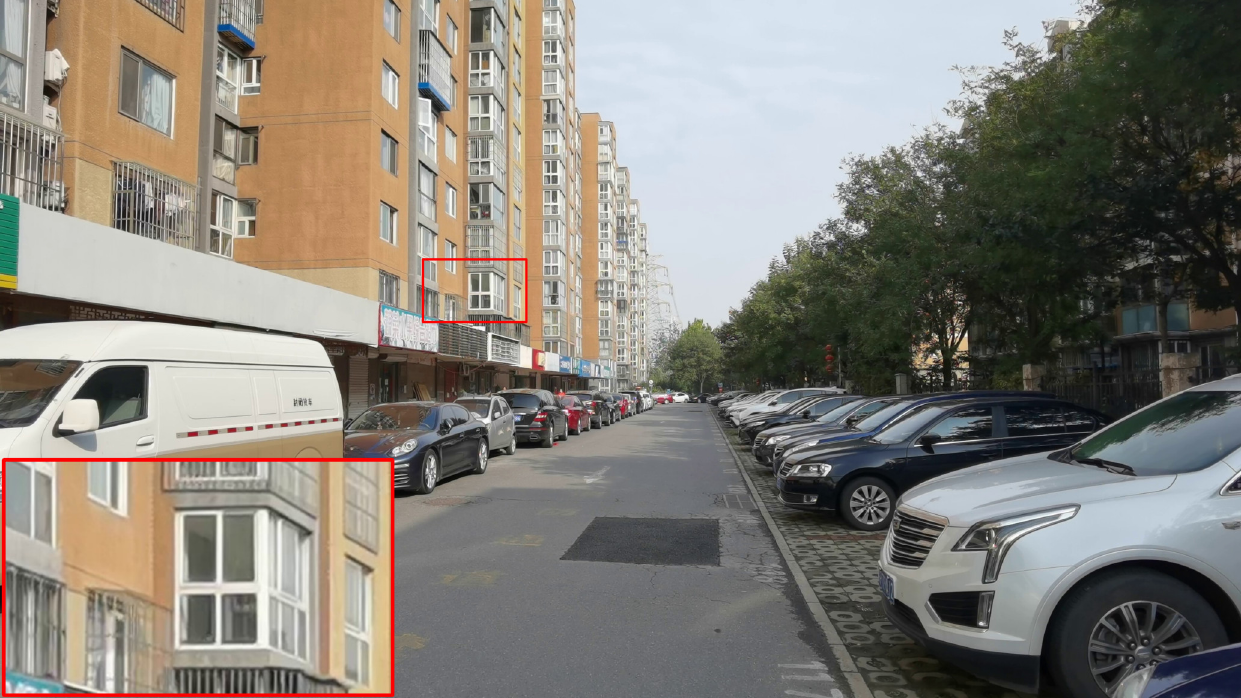} \\

    \includegraphics[width=0.16\textwidth]{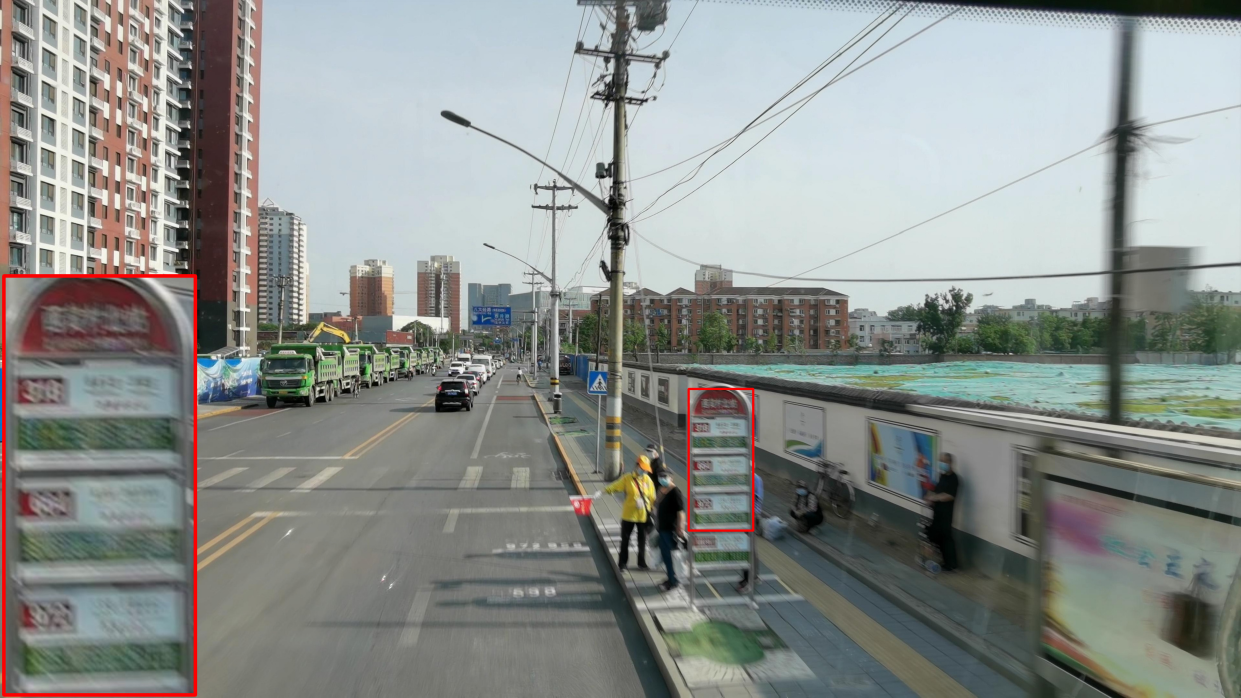} &
    \includegraphics[width=0.16\textwidth]{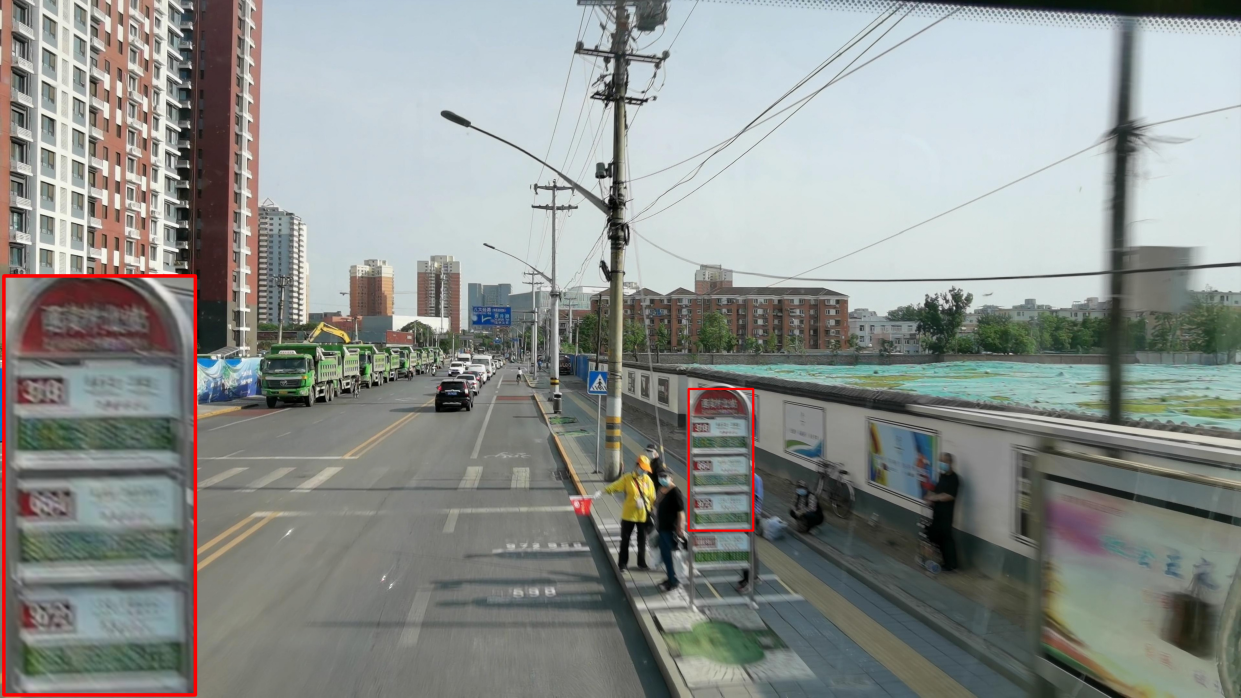} &
    \includegraphics[width=0.16\textwidth]{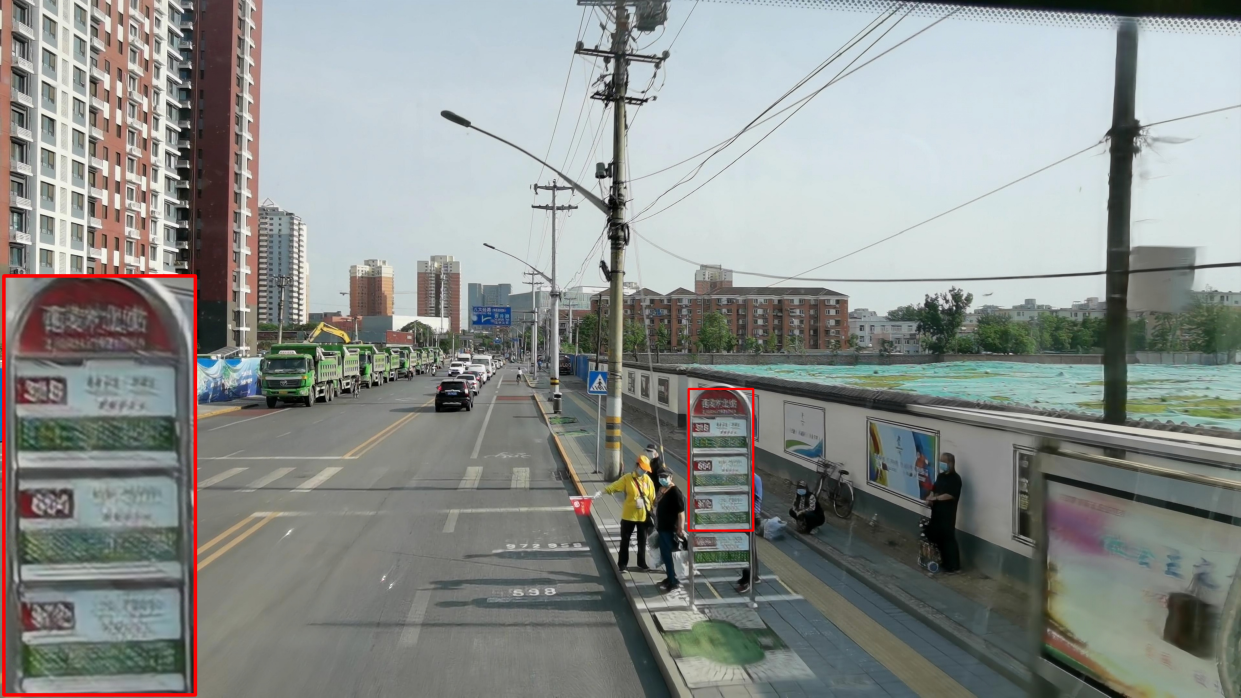} &
    \includegraphics[width=0.16\textwidth]{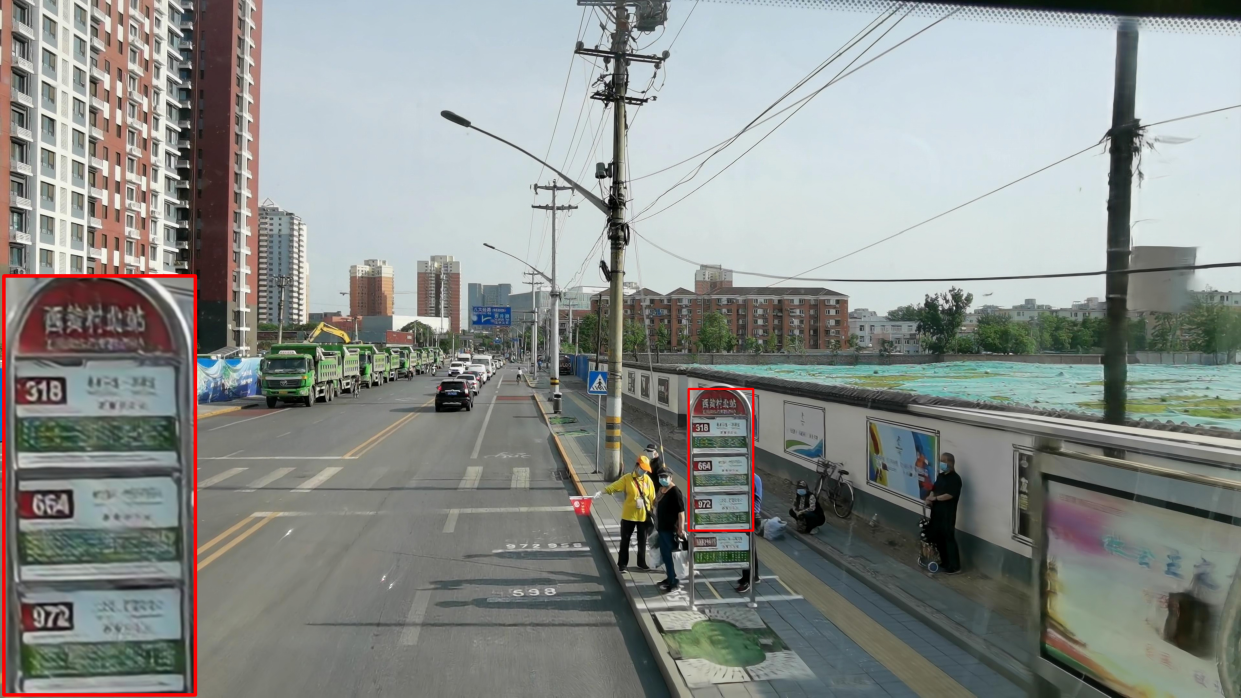} &
    \includegraphics[width=0.16\textwidth]{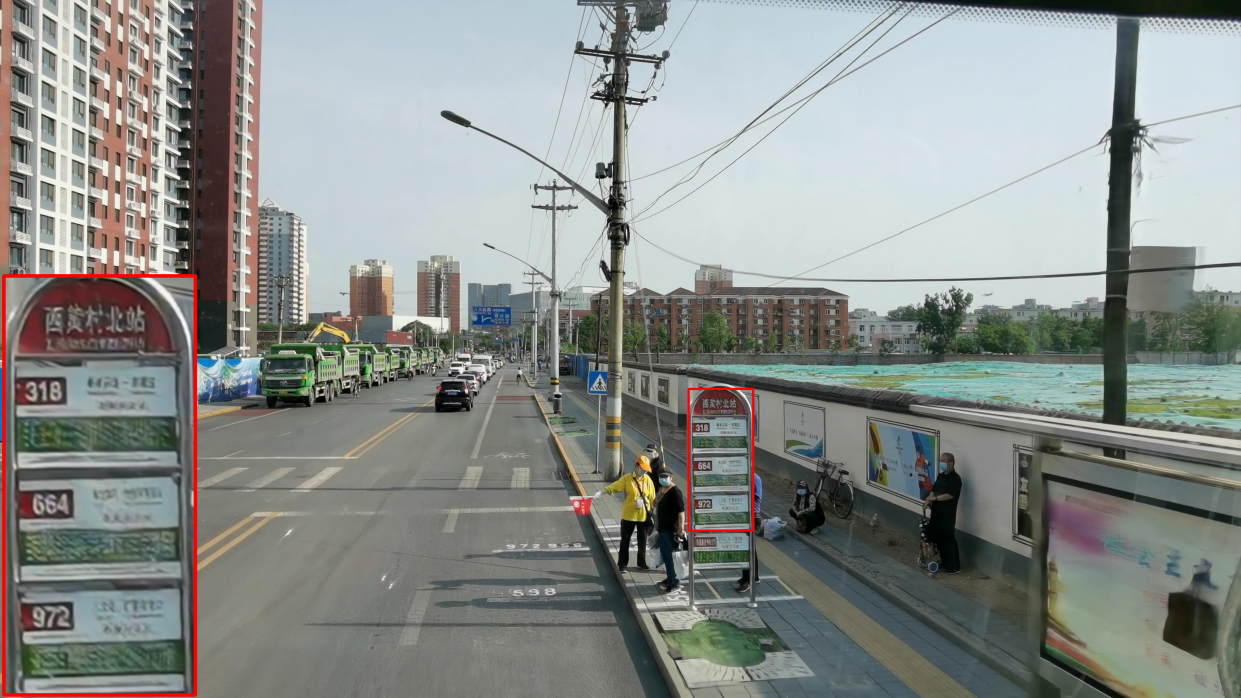} &
    \includegraphics[width=0.16\textwidth]{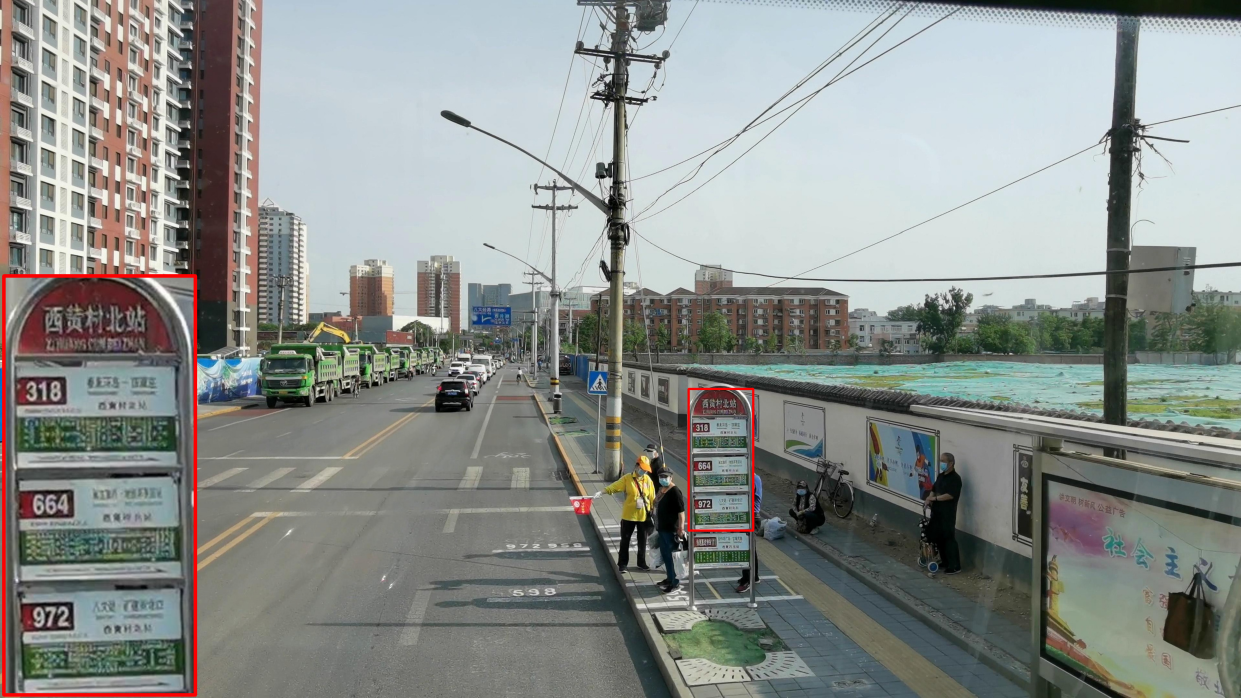} \\
    Input & UHDForm & ERR & D2Net & Ours & GT \\

  \end{tabular}

  \caption{Visual comparison between different architectures on Haze and Blur datasets \cite{wang2024uhdformer}. The first row shows results from UHD-Haze test (39,124) while the second row shows results from UHD-Blur test (3,147).}
  \label{fig:VisualComparison2}
\end{figure*}

\begin{figure}[ht]
\raggedright
\begin{minipage}{0.47\textwidth}
\raggedright                     
\captionof{table}{UHD-LL Quantitative \\Comparison}
\label{tab:LLQuantitative}
{\fontsize{8}{10}\selectfont
\begin{tabular}{l|c|c|c}
\hline
\textbf{Methods} & \textbf{PSNR$\uparrow$} & \textbf{SSIM$\uparrow$} & \textbf{Param.$\downarrow$} \\
\hline
IFT\cite{zhao2021deep} & 21.96 & 0.870 & 11.56M \\
SNR-Aware\cite{xu2022snr} & 22.72 & 0.877 & 40.08M \\
Uformer\cite{wang2022uformer} & 19.28 & 0.849 & 20.62M \\
Restormer\cite{zamir2022restormer} & 22.25 & 0.871 & 26.11M \\
DiffLL\cite{jiang2023low} & 21.36 & 0.872 & 17.29M \\
LLFormer\cite{Wang_Zhang_Shen_Luo_Stenger_Lu_2023} & 22.79 & 0.853 & 13.15M \\
UHDFour\cite{Li2023ICLR} & 26.23 & 0.900 & 17.54M \\
Wave-Mamba\cite{zou2024wavemamba} & 27.35 & 0.913 & 1.258M \\
UHDFormer\cite{wang2024uhdformer} & 27.11 & 0.927 & 0.339M \\
ERR\cite{Zhao_2025_CVPR} & \underline{27.57} & \underline{0.932} & 1.131M \\
\hline
\textbf{Ours} & \textbf{28.79} & \textbf{0.934} & 4.726M \\
\hline
\end{tabular}
}
\end{minipage}%
\hspace{0.3cm} 
\begin{minipage}{0.47\textwidth}
\raggedleft                      
\captionof{table}{4K-Rain13k Quantitative Comparison}
\label{tab:RainQuantitative}
{\fontsize{8}{10}\selectfont
\begin{tabular}{l|c|c|c}
\hline
\textbf{Methods} & \textbf{PSNR$\uparrow$} & \textbf{SSIM$\uparrow$} & \textbf{Param.$\downarrow$} \\
\hline
JORDER-E\cite{yang2019joint} & 30.46 & 0.912 & 4.21M \\
RCDNet\cite{wang2020model} & 30.83 & 0.921 & 3.17M \\
SPDNet\cite{9710307} & 31.81 & 0.922 & 3.04M \\
IDT\cite{xiao2022image} & 32.91 & 0.948 & 16.41M \\
Restormer\cite{zamir2022restormer} & 33.02 & 0.933 & 26.12M \\
DRSformer\cite{chen2023learning} & 32.94 & 0.933 & 33.65M \\
UDR-S2Former\cite{chen2023sparse} & 33.36 & 0.946 & 8.53M \\
UDR-Mixer\cite{chen2024towards} & 34.28 & 0.951 & 8.53M \\
ERR\cite{Zhao_2025_CVPR} & \underline{34.48} & \underline{0.952} & 1.131M \\
\hline
\textbf{Ours} & \textbf{34.50} & \textbf{0.967} & 4.726M \\
\hline
\end{tabular}
}
\end{minipage}
\begin{minipage}{0.47\textwidth}
\raggedright
\captionof{table}{UHD-Haze Quantitative Comparison}
\label{tab:HazeQuantitative}
{\fontsize{8}{10}\selectfont
\begin{tabular}{l|c|c|c}
\hline
\textbf{Methods} & \textbf{PSNR$\uparrow$} & \textbf{SSIM$\uparrow$} & \textbf{Param.$\downarrow$} \\
\hline
UHD\cite{zheng2021ultra} & 18.04 & 0.811 & 34.5M \\
Restormer\cite{zamir2022restormer} & 12.72 & 0.693 & 26.1M \\
Uformer\cite{wang2022uformer} & 19.83 & 0.737 & 20.6M \\
DehazeFormer\cite{song2023vision} & 15.37 & 0.725 & 2.5M \\
UHDFormer\cite{wang2024uhdformer} & 22.59 & 0.943 & 0.339M \\
D2Net\cite{wu2025dropout} & 24.88 & 0.944 & 5.22M \\
ERR\cite{Zhao_2025_CVPR} & \underline{25.12} & \underline{0.950} & 1.131M \\
\hline
\textbf{Ours} & \textbf{26.63} & \textbf{0.956} & 4.726M \\
\hline
\end{tabular}
}
\end{minipage}
\hspace{0.3cm}
\begin{minipage}{0.47\textwidth}
\raggedleft
\captionof{table}{UHD-Blur Quantitative\\ Comparison}
\label{tab:BlurQuantitative}
{\fontsize{8}{10}\selectfont
\begin{tabular}{l|c|c|c}
\hline
\textbf{Methods} & \textbf{PSNR$\uparrow$} & \textbf{SSIM$\uparrow$} & \textbf{Param.$\downarrow$} \\
\hline
MIMO-Unet++\cite{cho2021rethinking} & 25.03 & 0.752 & 16.1M \\
Restormer\cite{zamir2022restormer} & 25.21 & 0.752 & 26.1M \\
Uformer\cite{wang2022uformer} & 25.27 & 0.752 & 20.6M \\
Stripformer\cite{Tsai2022Stripformer} & 25.05 & 0.750 & 19.7M \\
FFTformer\cite{kong2023efficient} & 25.41 & 0.757 & 16.6M \\
UHDFormer\cite{wang2024uhdformer} & 28.82 & 0.844 & 0.339M \\
ERR\cite{Zhao_2025_CVPR} & 29.72 & 0.861 & 1.131M \\
D2Net\cite{wu2025dropout} & \underline{30.46} & \underline{0.872} & 5.22M \\
\hline
\textbf{Ours} & \textbf{30.71} & \textbf{0.886} & 4.726M \\
\hline
\end{tabular}
}
\end{minipage}
\end{figure}

\subsection{Quantitative and Qualitative Comparison}
\textbf{Low-Light Image Enhancement Results.} A comparison can be found in Table \ref{tab:LLQuantitative} between RetinexDual and SOTA methods on the \textbf{UHD-LL} dataset. Among the compared methods, IFT \cite{zhao2021deep}, SNR-Aware \cite{xu2022snr}, Uformer \cite{wang2022uformer}, Restormer \cite{zamir2022restormer}, DiffLL \cite{jiang2023low}, and LLFormer \cite{Wang_Zhang_Shen_Luo_Stenger_Lu_2023} perform poorly in both PSNR and SSIM. The remaining methods, such as UHDFour \cite{Li2023ICLR}, Wave-Mamba \cite{zou2024wavemamba}, UHDFormer \cite{wang2024uhdformer}, ERR \cite{Zhao_2025_CVPR}, and RetinexDual demonstrate more effectiveness for the task. While our model outperforms the previous SOTA method by 4.7\% in PSNR, RetinexDual achieves more SSIM than ERR by 0.2\%. Qualitative comparisons on challenging scenes show that RetinexDual provides the best visual results, as shown in Fig. \ref{fig:VisualComparison1}. 

\textbf{Image Deraining Results.} We train the proposed RetinexDual architecture on the \textbf{4K-Rain13K} dataset for the deraining task. We compare RetinexDual against recent models in the literature, inlcuding JORDER-E \cite{yang2019joint}, RCDNet \cite{wang2020model}, SPDNet \cite{9710307}, IDT \cite{xiao2022image}, Restormer \cite{zamir2022restormer}, DRSformer \cite{chen2023learning}, UDR-S2Former \cite{chen2023sparse}, UDR-Mixer \cite{chen2024towards}, and ERR \cite{Zhao_2025_CVPR}. Quantitative results in Table \ref{tab:RainQuantitative} show slight differences in terms of PSNR and SSIM between UDR-Mixer, ERR, and our model. However, RetinexDual still achieved the highest score among all models, in addition to delivering the best visual quality, especially in a challenging scene as shown in Fig. \ref{fig:VisualComparison1}. It is worth noting that DRSformer appears to be slightly worse than our model; however, RetinexDual achieved better PSNR and SSIM scores with roughly just 14\% of DRSformer's model size.

\textbf{Image Dehazing Results.} We evaluate the dehazing performance of RetinexDual on \textbf{UHD-Haze}. Table \ref{tab:HazeQuantitative} summarizes the results between our model and six other approaches, including UHD \cite{zheng2021ultra}, Restormer \cite{zamir2022restormer}, Uformer \cite{wang2022uformer}, DehazeFormer \cite{song2023vision}, UHDFormer \cite{wang2024uhdformer}, D2Net \cite{wu2025dropout}, and ERR \cite{Zhao_2025_CVPR}. While RetinexDual significantly outperforms SOTA methods by 6\% in terms of PSNR, and  0.06 in terms of SSIM. The visual results of the challenging samples show that RetinexDual significantly excels in producing perceptually clearer images, as in Fig. \ref{fig:VisualComparison2}.

\textbf{Image Deblurring Results.} Table \ref{tab:BlurQuantitative} summarizes the quantitative results on the \textbf{UHD-Blur} dataset, comparing RetinexDual with recent deblurring approaches such as MIMO-Unet++ \cite{cho2021rethinking}, Restormer \cite{zamir2022restormer}, Uformer \cite{wang2022uformer}, Stripformer \cite{Tsai2022Stripformer}, FFTformer \cite{kong2023efficient}, UHDFormer \cite{wang2024uhdformer}, ERR \cite{Zhao_2025_CVPR}, and D2Net \cite{wu2025dropout}. RetinexDual surpasses the SOTA performance, achieving 0.25 dB PSNR and 2\% SSIM improvements. Furthermore, our model consistently outperformed previous SOTA models perceptually when evaluated on high-distorted images, as shown in Fig. \ref{fig:VisualComparison2} shows how different models perform on a sample image from the UHD-Blur test set.

\subsection{Ablation Studies}

The ablation experiments are divided into two parts: the first one examines the effect of the constructed training objective function. The second validates the effectiveness of our contributions, structured into three parts, each addressing one contribution along with the relevant components in our model. We conducted our ablation studies on the \textbf{UHD-LL} dataset with visual comparisons and a detailed description of each study in the Supplementary material.
\begin{table}
\caption{Ablation study for training objective}
\centering
{\fontsize{7.5}{9.5}\selectfont
\begin{tabular}{@{\hskip 3pt}c|c@{\hskip 2pt}c@{\hskip 2pt}c@{\hskip 2pt}c|@{\hskip 2pt}c@{\hskip 2pt}|@{\hskip 2pt}c@{\hskip 2pt}||@{\hskip 2pt}c@{\hskip 2pt}|@{\hskip 2pt}c@{\hskip 3pt}}
\hline
\textbf{Exp.} & \textbf{$\mathcal{L}_{cb}$} & \textbf{$\mathcal{L}_{fft}$} & \textbf{$\mathcal{L}_{ssim}$}  & \textbf{$\mathcal{L}_{p}$} & \textbf{Multilevel} & \textbf{Scaling} & \textbf{PSNR$\uparrow$} & \textbf{SSIM$\uparrow$} \\
\hline
1  & \ding{55} & \ding{51} & \ding{51} & \ding{51} & \ding{51} &  \ding{51} & 27.27 & 0.911 \\ 
2  & \ding{51} & \ding{55} & \ding{51} & \ding{51} & \ding{51} &  \ding{51} & 27.92 & 0.918 \\ 
3  & \ding{51} & \ding{51} & \ding{55} & \ding{51} & \ding{51} &  \ding{51} & 27.15 & 0.906 \\ 
4  & \ding{51} & \ding{51} & \ding{51} & \ding{55} & \ding{51} &  \ding{51} & 27.20 & 0.911 \\ 
5  & \ding{51} & \ding{51} & \ding{51} & \ding{51} & \ding{55} &  \ding{55} & 26.63 & 0.906 \\  
6  & \ding{51} & \ding{51} & \ding{51} & \ding{51} & \ding{51} &  \ding{55} & 28.54 & 0.921 \\ 
\hline
\textbf{7} & \ding{51} & \ding{51} & \ding{51} & \ding{51} & \ding{51} &  \ding{51} & \textbf{28.79} & \textbf{0.934} \\
\hline
\end{tabular}
}
\label{tab:Ablation_Losses}
\end{table}
\begin{table}
\caption{Ablation study for key contributions}

\centering
{\fontsize{7.5}{9.5}\selectfont
\begin{tabular}{l|l|c|c|c}
\hline
\multicolumn{2}{l|@{\hskip 2pt}}{\textbf{Method}} & \textbf{PSNR$\uparrow$} &\textbf{SSIM$\uparrow$} & \textbf{Param.$\downarrow$} \\

\hline
\multicolumn{5}{c}{\textbf{RetinexDuality}}\\
\hline
A & w/ FIA for $R \& L$                  & 24.83 & 0.866  & 0.419M \\ 
B & w/ SAMBA for $R \& L$                & 27.91 & 0.920  & 9.032M \\ 
C & w/o FIA branch                       & 26.67 & 0.919  & 4.516M \\ 
D & w/o SAMBA branch                     & 23.83 & 0.866  & 0.209M \\ 
\hline
\multicolumn{5}{c}{\textbf{Scale Attentive maMBA (SAMBA)}}\\
\hline
A & w/o Multi-scale                      & 26.96 & 0.906  & 1.335M \\ 
B & w/o GSSB                            & 25.49 & 0.908  & 1.704M \\ 
C & w/o GSSB \& Multi-scale             & 24.83 & 0.876  & 0.904M \\ 
\hline
\multicolumn{5}{c}{\textbf{Frquency Illumination Adaptor (FIA)}}\\
\hline
A & w/o Fourier            & 27.03 & 0.919  & 4.682M \\ 
\hline
\hline
\multicolumn{2}{l|}{\textbf{Full Model (ours)}}          & \textbf{28.79} & \textbf{0.934}  & 4.726M \\
\hline
\end{tabular}
}
\label{tab:Ablation_Archs_Contributions}
\end{table}

\textbf{Ablation Studies on the Training Objective.} We report the results of seven experiments in Table \ref{tab:Ablation_Losses}, where the last experiment represents our full model without modifications. It was noticed that by removing the multi-level loss, there was a huge drop in performance in the model, which confirms its contribution. Among all experiments, experiment 7 records the highest PSNR and SSIM values, indicating the importance of having such a training objective function.

\textbf{Ablation Studies of Key Contributions.} We conduct the key contributions ablation study in three parts, as shown in Table \ref{tab:Ablation_Archs_Contributions}. Each of our contributions is validated through multiple experiments that demonstrate its significance. The first set of experiments confirms the importance of having two different sub-networks to process the two components of the image. It can be noticed that even by using the more complex branch (SAMBA) for both components, the approach doesn't reach higher performance. The second set, on the other hand, demonstrates the effectiveness of using not only Mamba in the reflectance branch, but also having multiple scales of the feature maps inside it, which is confirmed by removing those parts. Finally, the results obtained by the third set support that correcting the illumination component in the frequency domain is fit-for-purpose.

\begin{table}
\caption{Comparison of inference time on 4K images}
\centering
{\fontsize{7.5}{9.5}\selectfont
\begin{tabular}{l|c@{\hskip 7pt}c@{\hskip 7pt}c@{\hskip 7pt}c@{\hskip 7pt}c}
\hline
\textbf{Method}                 & Wave-Mamba    &   ERR     &   D2Net   & Ours \\
\hline
Inference Time (s)$\downarrow$  & 0.957         &  0.601    &   1.63    & 0.955 \\
\hline
\end{tabular}
}
\label{tab:InferenceTimeComparison}
\end{table}

\section{Conclusion}
In this paper, we have developed RetinexDual, which is a novel UHD image restorer based on the Retinex theory. RetinexDual decomposes the image into reflectance and illumination components, each handled using a distinct sub-network designed to suit its specific characteristics. Evaluating our model on four different UHD datasets demonstrates that it outperforms the SOTA methods both qualitatively and in terms of PSNR and SSIM. We have also presented ablation studies that validate the contribution of each component in RetinexDual. Although our approach achieved desirable results, it is not optimized in terms of inference time and size, as shown in Table \ref{tab:InferenceTimeComparison}. Additionally, like earlier models, our model faces some difficulty in restoring certain challenging images. Thus, there is still room for improvement in future work.

\bibliographystyle{splncs04}
\bibliography{mybibliography}

\end{document}